%% file: main.tex
\documentclass[letterpaper]{article} %
\usepackage{aaai2026}  %
\usepackage{times}  %
\usepackage{helvet}  %
\usepackage{courier}  %
\usepackage[hyphens]{url}  %
\usepackage{graphicx} %
\usepackage{natbib}  %
\usepackage{caption} %
\usepackage{algorithm}
\usepackage{algorithmic}
\usepackage{tikz}
\usetikzlibrary{
arrows.meta,
decorations.pathmorphing,
patterns,
shapes.geometric,
shapes.symbols,
calc
}
\usepackage{xcolor}
\usepackage{newfloat}
\usepackage{listings}
\DeclareCaptionStyle{ruled}{labelfont=normalfont,labelsep=colon,strut=off} %
\floatstyle{ruled}
\newfloat{listing}{tb}{lst}{}
\floatname{listing}{Listing}
\usepackage{booktabs}
\usepackage{tabularx}
\usepackage{amssymb}
\usepackage{makecell}

\title{Requirements for Aligned, Dynamic Resolution of \\ Conflicts in Operational Constraints}
\author{
    Steven J. Jones, 
    Robert E. Wray, 
    John Laird 
}
\affiliations{
    Center for Integrated Cognition at IQMRI\\

    Ann Arbor, MI 48105 USA\\
    \{steven.jones,robert.wray,john.laird\}@cic.iqmri.org
}

\nocopyright

\usepackage{xcolor}

\newif\ifshowcomments
\showcommentstrue %

\begin{document}

\maketitle

\begin{abstract}
Deployed, autonomous AI systems must often evaluate multiple plausible courses of action (extended sequences of behavior) in novel or under-specified contexts. Despite extensive training, these systems will inevitably encounter scenarios where no available course of action fully satisfies all operational constraints (e.g., operating procedures, rules, laws, norms, and goals). To achieve goals in accordance with human expectations and values, agents must go beyond their trained policies and instead construct, evaluate, and justify candidate courses of action. These processes require contextual ``knowledge'' that may lie outside prior (policy) training. This paper characterizes requirements for agent decision making in these contexts. It also identifies the types of knowledge agents require to make decisions robust to agent goals and aligned with human expectations. Drawing on both analysis and empirical case studies, we examine how agents need to integrate normative, pragmatic, and situational understanding to select and then to pursue more aligned courses of action in complex, real-world environments.
\end{abstract}

\begin{links}\small
    \link{Code and Technical Appendix}{https://github.com/Center-for-Integrated-Cognition/OAMNCC}
\end{links}

\section{Introduction}

In open environments, agents will face conflicting goals and constraints \cite{freuder_partial_1992}, implicit expectations \cite{shah_preferences_2019}, and incomplete information \cite{kaelbling_planning_1998}. However, the behavior of autonomous agents in these open environments should remain \textit{aligned}: their behavior must appear reasonable and appropriate to human observers and be consistent with their expectations, laws, norms, etc.

Alignment in open environments cannot be wholly solved with \textit{a priori} training; agents will face difficulty pursuing their intended objectives in specific, dynamic, and often unfamiliar/novel environments \cite{langosco_goal_2022}. This difficulty, and the benefits of justified deviations from external constraints, indicate a need to go beyond static preference satisfaction or fixed rules to more dynamic models of alignment \cite{milli_should_2017}.%

We emphasize one aspect of such alignment: conforming to constraints. In real-world environments, autonomous agents are subject to a dense network of constraints: formal regulations and laws, social norms, organizational rules (e.g., military doctrine), and task-specific guidelines and preferences. These constraints are often ambiguous, underspecified, or even contradictory in practice, especially when instantiated in complex or dynamic environments \cite{wray_computational-level_2023}.
Human decision-making in such contexts relies on interpretive flexibility, sensitivity to context, and the ability to weigh competing considerations \cite{klein_sources_1998,payne_adaptive_1993}. Similarly, autonomous systems must be able to evaluate whether their overall course of action is appropriate given the situational constraints and tradeoffs involved. Eliminating all conflicts through training or rule design alone is not feasibile: agents must revise their commitments and plans in response to dynamic contexts \cite{bratman_intentions_1987}, which include  large, open-ended, and novel combinations and interactions among constraints.

Our initial research toward online conflict mitigation clarified the vast scope and complexity of agent knowledge needed for comprehensive mitigation \cite{jones_toward_2024_correct,wray_computational-level_2023}. Here, we step back from evaluating specific, implemented solutions and characterize the problem and its solution space more completely. We adopt a knowledge-level perspective \cite{newell_knowledge_1982}. Rather than specify how an agent  should respond when facing novel constraint conflicts (i.e., ones not anticipated pre-deployment, such as pre-training), we identify types of knowledge an agent must possess for identifying and selecting aligned courses of action.

We refer to ``knowledge'' in the broad, functional sense of Newell's ``knowledge level.'' The analysis treats knowledge as what an agent must be able to represent and do to achieve specific outcomes in a conflict, without committing to underlying realizations (e.g.,  symbolic vs.  distributed representations), and regardless of whether the knowledge is procedural or declarative. The analysis identifies specific agent capacities (such as the expressivity of representations, contextual sensitivity, and inferential capacities) relevant to achieving operationally successful and aligned behavior. 

Beyond pre-learned rules or preference hierarchies, we argue that such knowledge must include interpretive resources such as frames for understanding different kinds of norms, background expectations about tradeoffs and exceptions, and the ability to assess ``reasonable'' deviation from a constraint. The knowledge-level analysis aims to establish general requirements, regardless of an agent's implementation, to navigate conflicts in a manner that aligns with human expectations and values in similar decision-making contexts.

\begin{table*}[t]
    \centering
    {%
    \begin{tabularx}{.975\textwidth}{l X} 
        \toprule
        \textbf{Conflict Category} & \textbf{Cause of the Conflict} \\
        \midrule
        
        \textbf{Unsatisfiability} & Two or more constraints' logical entailments specify contradictory prescriptions on behavior. \\
        \hspace*{1em} Infeasibility & Two constraints specify prescriptions on behavior that are directly contradictory. \\
        \hspace*{1em} Incommensurability & Constraints appeal to value systems without a common unit of measure for compliance. \\
        
        \textbf{Mutual Exclusivity} & Constraints may be coherent, but available actions do not afford complete compliance. \\
        \hspace*{1em} Resource Contention & Two or more required actions compete for the same insufficient finite resource. \\
        \hspace*{2em} \textit{Temporal} & The finite resource is time. \\
        \hspace*{2em} \textit{Spatial} & The finite resource is space. \\
        \hspace*{1em} Causal Preclusion & Action to satisfy one constraint changes the world, making another constraint impossible. \\

        \textbf{Uncertainty} & A conflict arises or is made worse by a lack of knowledge of the state or expectations. \\
        \hspace*{1em} Epistemic & The agent lacks knowledge about the world. \\
        \hspace*{1em} Probabilistic & The world is stochastic, so outcomes cannot be determined. \\
        
        \bottomrule
    \end{tabularx}}
    \caption{A Taxonomy of Constraint Conflict Types}
    \label{tab:conflict_taxonomy}
\end{table*}

The paper develops this knowledge-level perspective by enumerating types of constraint conflicts and illustrating (via representative scenarios) how different knowledge sources support conflict-resolution behavior. The scenarios highlight information agents must access to construct and justify appropriate courses of action. These examples ground proposed knowledge requirements that can inform both design and evaluation of future systems. By clarifying the kinds of knowledge needed to navigate constraint conflicts in open contexts, this work provides a foundation for the development of future agents whose behavior will remain coherent, context-sensitive, and aligned, even in environments beyond prior specification.%

\section{Problem Definition and Initial Analysis}
\label{section:initial-analysis}

\begin{algorithm}[tb]
\caption{Schematic process model for \underline{O}nline, \underline{A}ligned \underline{M}itigation of \underline{N}ovel \underline{C}onstraint \underline{C}onflicts (OAMNCC).}
\label{alg:bare-process-model}
\begin{algorithmic}[1] %
\STATE Recognize novelty of the conflict \\ (similar to ``out of distribution'' assessment)
\STATE Assess conflict type and structure  
\STATE Evaluate what situational information might be relevant to the conflict (expand space of consideration)
\STATE Propose candidate conflict-mitigating courses of action, evaluate them, and select one
\STATE Execute the selected course of action, monitoring for resolution of conflict
\end{algorithmic}
\end{algorithm}

Algorithm~\ref{alg:bare-process-model} outlines the processing an agent might perform to mitigate conflicts in novel conflict situations. The  outline is based on aforementioned early prototypes of such a capability and reflects high-level commitments similar to those of Goal-Driven Autonomy \cite{molineaux_goal-driven_2010}. Such processing must occur in the performance environment (\textit{online}), and the outcomes it produces should be aligned with human expectations. For brevity of presentation, we use OAMNCC as a shorthand for this capability. 

In Step~1 of the process, an autonomous agent must recognize novel conflicts. Mechanisms for such recognition are comparable to out-of-distribution detection in learning systems. However, for mitigation, as for learning, detection must occur \emph{before} taking action using existing policy \cite{haider_can_2024}. Step~2 reflects an ability to characterize the detected problem among constraints that needs solving. In Steps~3 and~4, an agent  weighs what information in its context is relevant and generates candidates that exploit newly-identified information. (As we discuss below, these steps require further specification, which we discuss in the following sections.) Using this algorithm as a scaffold, we seek to identify what knowledge is required to enable OAMNCC in an agent. %

Initial investigations identified a few abstract requirements \cite[similar to those identified by others; e.g.,][]{kuipers_how_2018}. Agents must demonstrate awareness of the frames by which humans interpret constraints (e.g., duty, norms, utility). Agents must also be able to represent preferences for responding to conflicts with a similar level of expressivity to human preferences. A constraint-compliant agent must also represent: its situation, how constraints bear on its behavior, its available courses of action, and what those actions can be expected to achieve (or, minimally, whether they can be expected to violate constraints). %
All of these kinds of knowledge are necessary. If any one is missing, compliance is infeasible. When all are present, an agent can feasibly select courses of action in its current situation that it expects will comply with constraints \cite{wray_computational-level_2023,rao_bdi_1995}.

While useful, these requirements are incomplete, given the overall space of constraint conflicts. Table~\ref{tab:conflict_taxonomy} summarizes the different types of conflicts relevant to conflict mitigation. The table spotlights the (very large) scope of the problem. Any mitigation solution must span all of these different types of conflicts. We use this list to assess coverage of the knowledge-level analysis over the entirety of the mitigation challenge. The list also hints at the knowledge needed to assess conflicts (i.e., perform Step~2).

To motivate the taxonomy, we construe conflicts as arising when the knowledge content of at least one of the representations outlined previously is insufficient for a constraint-compliant choice. Thus, the major types of conflicts result from different kinds of knowledge inadequacy. When representations of how constraints bear on behavior entail contradictory or involve incommensurable values, the agent faces an unsatisfiable conflict of constraint semantics. When the constraints are coherent, but the agent's representations of its situation, actions, and their expected outcomes reveal no fully compliant course of action, the agent faces a situational conflict. Situational conflicts can arise from mutual exclusivity or uncertainty (such as from abstract constraint specification or partial observability). Finally, multiple, different conflict types may manifest in a given conflict situation.

We observe that some conflict types have been more thoroughly investigated than others, and some are more amenable to semantic analysis and pretraining. For example, techniques have been identified to mitigate unsatisfiability conflicts at the time the constraint specifications are introduced, especially rule-oriented constraints \cite{censi_liability_2019}. This enables offline, pretraining approaches to resolve such conflicts. Generally, conflict types inherent to the definition and expression of constraints themselves can be more readily avoided via preparatory strategies such as training. We thus focus on situational conflicts. These conflicts involve mutual exclusivity and uncertainty from interactions between constraints and the open-world situation, where it is not feasible to anticipate all possible interactions in advance. %

\section{Example Constraint Conflict Scenarios}%

We introduce scenarios designed to illustrate additional knowledge needed for timely, coherent conflict mitigation that is aligned with human expectations (OAMNCC). %

The scenarios here build on ones previously developed by researchers at the Naval Postgraduate School  \cite[NPS;][]{brutzman_ethical_2020}.\footnote{Specific parameters for general capabilities, speed, and fuel consumption were adopted from public data for appropriate ship types. 
Events such as the pirate attacks and sailor rescue are modeled with simple probability distributions reflecting estimates from available data.}%
In these scenarios, an agent ``commands'' an autonomous surface vessel, a role comparable to a ship's captain, in a naval domain where dynamic conflicts are likely to arise. The original scenarios do not explicitly detail constraint conflicts. However, the scenarios do describe abstract ``course of action'' policy knowledge designed to comply with real-world, operational constraints (e.g., rules of engagement, specific limited resources). We adapt these scenarios to illustrate specific types of conflicts.%

Each scenario introduces one type of conflict that could arise in that situation. We then interrogate what kinds of knowledge are required to resolve the scenario in a way that appears consistent with human expectations. These represent knowledge requirements. We note each individual requirement appearing  in each scenario and then discuss the requirements collectively in the following section. The approach is analytic: we did not conduct studies to determine specific decisions that humans would make, as others have done in explorations of aligned, online sequential decision-making tasks \cite[e.g.,][]{loreggia_making_2022}. However, these scenarios are sufficiently straightforward that it is readily evident how the information we identify is relevant to resolving the conflict. Further, failing to take this information into account leads to behavior and outcomes inconsistent with (likely) human expectations and values. %

Two further introductory assumptions:
\begin{itemize}
    \item OAMNCC proceeds as outlined in the process model introduced above. The agent has available a few different online decision-making strategies when novel conflicts arise. The specific decision strategies are not intended to be exhaustive or comprehensive. Rather, they are reasonable for these domains and sufficiently familiar to enable us to isolate the importance of knowledge that would be expected to contribute to agent decision making.
    \item Consistent with a knowledge-level framing, we assume relevant knowledge is available for decision making. We do not evaluate the differential impact of immediate accessibility (i.e., recall from memory or \textit{knowledge search}) vs.  search for the information (internal or external \textit{problem search}).  Knowledge and problem search strategies are critical to the quality and timeliness of decisions; future efforts to implement OAMNCC  will need to address both. 
\end{itemize} %
 
From a methodological perspective, we designed specific simulation scenarios (implemented in Python) to support the different uses cases described in the paper. Scenarios were designed with distributions over various parameters (e.g., initial positions of objects). Data were collected for each scenario instance across the distribution(s) relevant to that scenario. Aggregate statistics across these runs constitute the results presented below.

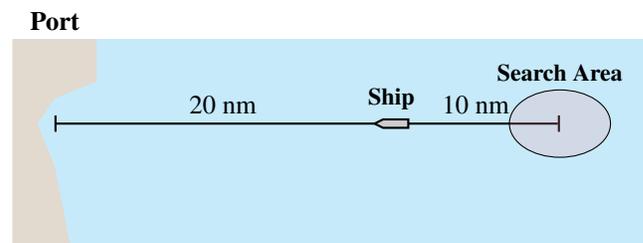
\begin{figure}[b]
    \centering
\begin{tikzpicture}[scale=0.112]
\fill[cyan!20] (-5,5) rectangle (70,30);
  \fill[brown!50!gray!30] (-5,5) -- (-5,30) -- (5,30) -- (5,25) -- (0,23) -- (-2,20) -- (0,15) -- (1,10) -- (2,5) -- cycle;
  \node[align=center, anchor=south west] at (-4,30) {\textbf{Port}};

  \coordinate (port) at (0,20);
  \coordinate (ship) at (40,20);
  \coordinate (searchCenter) at (60,20);

  \draw[|.-., thick] (port) -- (ship.west) node[midway, above, sloped] {20 nm};
  \draw[.-.|, thick] (ship) -- (searchCenter) node[midway, above, sloped] {10 nm};

  \draw[fill=gray!40, thick] 
    (38,20) -- (39,20.5) -- (42,20.5) -- (42,19.5) -- (39,19.5)--cycle;
  \node at (40,23) {\footnotesize \textbf{Ship}};

  \fill[red!40, opacity=0.2] (60,20) ellipse (6 and 4);
  \draw (60,20) ellipse (6 and 4);
  \node at (60,26) {\footnotesize\textbf{Search Area}};
\end{tikzpicture}
    \caption{Illustration of Sailor Overboard Scenario.}
    \label{fig:sailor-overboard}
\end{figure}

\subsection{Sailor Overboard}

The drone ship has an urgent/high-priority need to return to base (RTB). During transit, a sailor is lost overboard without alert. The loss is then discovered 20nm from the ship's destination, as illustrated in Figure~\ref{fig:sailor-overboard}. The ship backtracks and searches until fuel runs so low that returning to base is endangered. In the scenario, the sailor overboard occurs 25-35nm from port (5-15nm from the alert location). Ignoring differences in fuel consumption for transit vs. search-and-rescue activities, the ship can safely backtrack 10nm and still reach port before fuel is exhausted. The rescue operation itself is not instantaneous and requires time and fuel resources. 
\begin{figure}
    \centering
    \includegraphics[width=0.991\columnwidth]{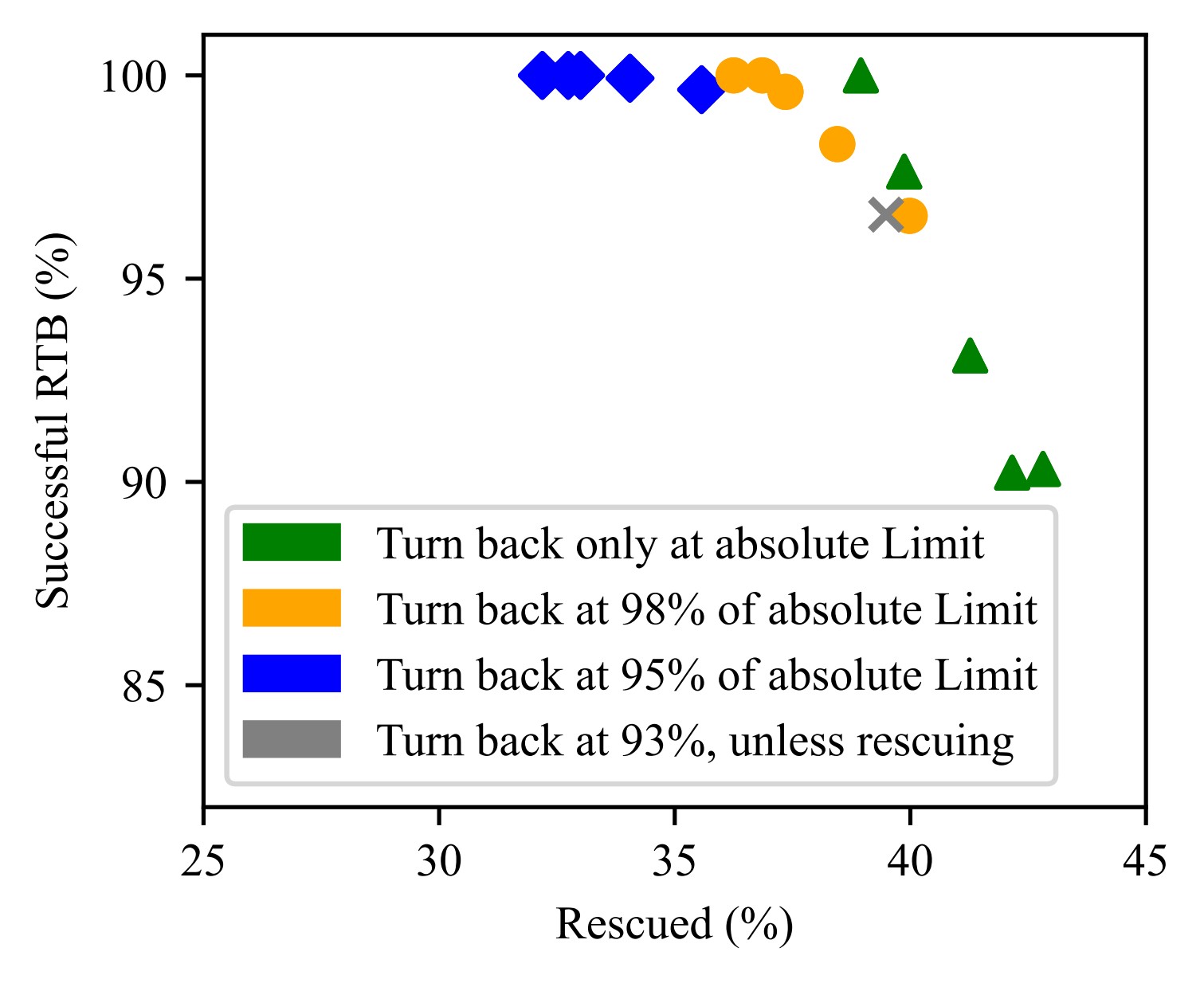}%
    \caption{Utilitarian assessment of sailor overboard scenario. Colored dots represent different safety margins. Identical marks vary in the ratio of rescue/RTB importance. $\times$ represents a policy that never leaves a sailor behind once spotted.}%
    \label{fig:overboard}
\end{figure}

A utilitarian decision frame can be used to assess the costs of returning to base vs. completing the rescue. Such a strategy can be tuned to achieve a desired evaluation of the tradeoff \cite{rossi_soft_2006}. Figure~\ref{fig:overboard} illustrates the tradeoff over 1000 simulated scenarios. The figure shows the relationship between successful rescues (x-axis) and successful RTBs (y-axis) as a function of two criteria: margin of error (different colors) and relative weighting of importance between rescue and RTB (different dots of the same color). For example, the 95\%-conservative policy (blue) curtails the number of rescues, regardless of the relative cost of rescue in comparison to RTB. 

From the figure, it is evident that a policy tuned to the ``knee'' of the curve prevents excessive rescue labor for diminishing returns and is also less sensitive to the specific, relative importance/cost of RTB vs. rescue, a value is  not likely to be known or explicit in this domain. Assuming the urgency and importance of the RTB order, this decision framing and knowledge at first appear an apt (or at least sufficient) decision strategy.

This presentation hides an important detail, however. This framing results, in some cases, with the vessel turning back \textit{after} having spotted a sailor. Such decisions contradict a norm to treat rescue as a duty once a rescue target is spotted \cite{mckie_rule_2003,jenni_explaining_1997}. Such a responsibility is a form of relevant information needed for Step~3 of Algorithm~\ref{alg:bare-process-model}. A policy that achieves nearly the same performance (i.e., a candidate response proposed in Step~4 of the process model, marked with $\times$ in Figure~\ref{fig:overboard}) can be attained by turning back to base either when returning to base is jeopardized (but a sailor is not spotted), or returning only after their rescue (once spotted). 

This scenario highlights an aspect of human preferences that needs to be reflected in agent knowledge: humans have preferences defined with respect to specific grounding statuses of constraints (Rq: \textit{expressive preferences}).%

\subsection{Piracy Interdiction}

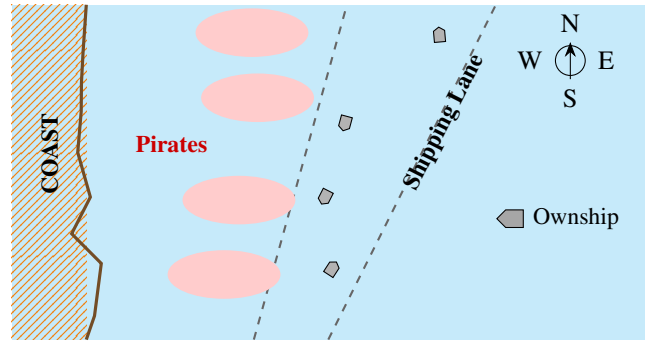
\begin{figure}[bt!]
    \centering
\begin{tikzpicture}[scale=0.495,
    ship/.style={
        draw,
        fill=gray!70,
        signal,
        signal to=west,
        minimum height=.2cm%
    },
    pirate/.style={
        ellipse,
        fill=red!20,
        minimum width=1.5cm,
        minimum height=.5cm,
        inner sep=6.5pt
    }
]

\fill[cyan!20] (-1,0) rectangle (16,9);

\fill[pattern=north east lines, pattern color=orange!80!brown] (-1,0) rectangle (1,9);
\draw[decoration={random steps, segment length=5mm, amplitude=2mm}, decorate, very thick, brown!60!black]
    (1,0) -- (1,9);
\node[rotate=90, anchor=south, font=\small] at (0.5, 5) {\textbf{COAST}};

\draw[dashed, thick, black!60] (5.5,0) -- (8,9);
\draw[dashed, thick, black!60] (7.5,0) -- (12,9);
\node[rotate=63, anchor=south, font=\small] at (11.1, 5.6) {\textbf{Shipping Lane}};

\node[pirate] at (4.7, 1.75) {};
\node[pirate] at (5.1, 3.75) {};
\node[pirate] at (5.6, 6.5) {};
\node[pirate] at (5.45, 8.25) {};
\node[red!80!black, font=\bfseries\small, anchor=east] at (4.5, 5.25) {Pirates};

\node[ship, label={[font=\small]right:Ownship}] at (12.5, 3.25) {};

\newcommand{\merchantship}[3]{
  \begin{scope}[shift={(#1,#2)}, rotate=#3]
    \draw[fill=gray!60]
      (0,.4) --          %
      (.15,.3) --
      (.15,0) -- %
      (-.15,0) -- %
      (-.15,.3) --
      cycle;
  \end{scope}
}

\merchantship{7.5}{1.75}{-33}   %
\merchantship{7.5}{4}{153}  %
\merchantship{8}{6}{165}  %
\merchantship{10.5}{8}{5}

\begin{scope}[shift={(14,7.5)}]
    \draw[-{Stealth[length=2mm]}] (0,-0.5) -- (0,0.5) node[above] {N};
    \draw (0,-0.5) node[below] {S};
    \draw (-0.5,0) node[left] {W};
    \draw (0.5,0) node[right] {E};
    \path[draw] (0,0) circle (0.4cm);
\end{scope}
\end{tikzpicture}
\caption{Illustration of the Piracy Interdiction Scenarios.}
\label{fig:pirate-interdiction}
\end{figure}

In the remaining scenarios, we envision a mission context in which the autonomous drone ship is responsible for protecting a shipping lane from pirates operating in small boats from coastal waters. In response to the effectiveness of the drone ship, the pirates are adopting a new strategy to mass multiple, coordinated attacks within the drone's area of operation (AO). In these scenarios, the pirates conduct four simultaneous attacks on four merchant vessels in shipping lanes within the drone's AO, as sketched in Figure~\ref{fig:pirate-interdiction}. By design, the agent will only be capable of attempting to interdict one attacking pirate band.

The simulation models pirate attacks with a 95\% success rate \cite{leontarakis_piracy_2015} and taking place for up to 30 minutes \cite{icc_international_maritime_bureau_piracy_2024}. For the results in this section, we simulated  1000  instances of the scenario. The main parameter in each instance was the initial location of merchant ships in the sea lane. As specified by NPS, the appropriate behavior in an instance of piracy interdiction involves multiple steps (identity friend and foe, force escalation, etc.). Thus, we assume that a response guaranteeing the safety of a selected merchant vessel consumes enough time that the drone ship, at full speed, could not reach another merchant within a 30-minute attack window.

The agent receives no advance warning of the pirate attacks.  Once within attack range, individual pirate attacks have a probability of successful boarding every 1 min of simulated time. Unimpeded attacks conclude with eventual pirate success 95\% of the time, or once a maximum engagement-time threshold is exceeded 5\% of the time. If the drone ship reaches the pirates before their attack succeeds, the simulation assumes a successful interdiction (i.e., pirate attack fails).

We model the agent's response to the conflict as the selection of one of four abstract courses of action, given the agent's available information. Typically, interdiction is a duty (standing order). In this case, the conflict occurs between four instances of the same duty. Notably, simple prioritization rules among constraints will not resolve the conflict. Additional information is required.
We assume the agent has accessible, relevant information  (i.e., in Step~3 of Algorithm~\ref{alg:bare-process-model}), including the approximate ransom values of targeted ships, the locations of targeted ships, and an estimate of pirate success (boarding) given the time it will take to reach each merchant vessel.

\begin{figure}[t]
    \centering
    \includegraphics[width=0.991\columnwidth]{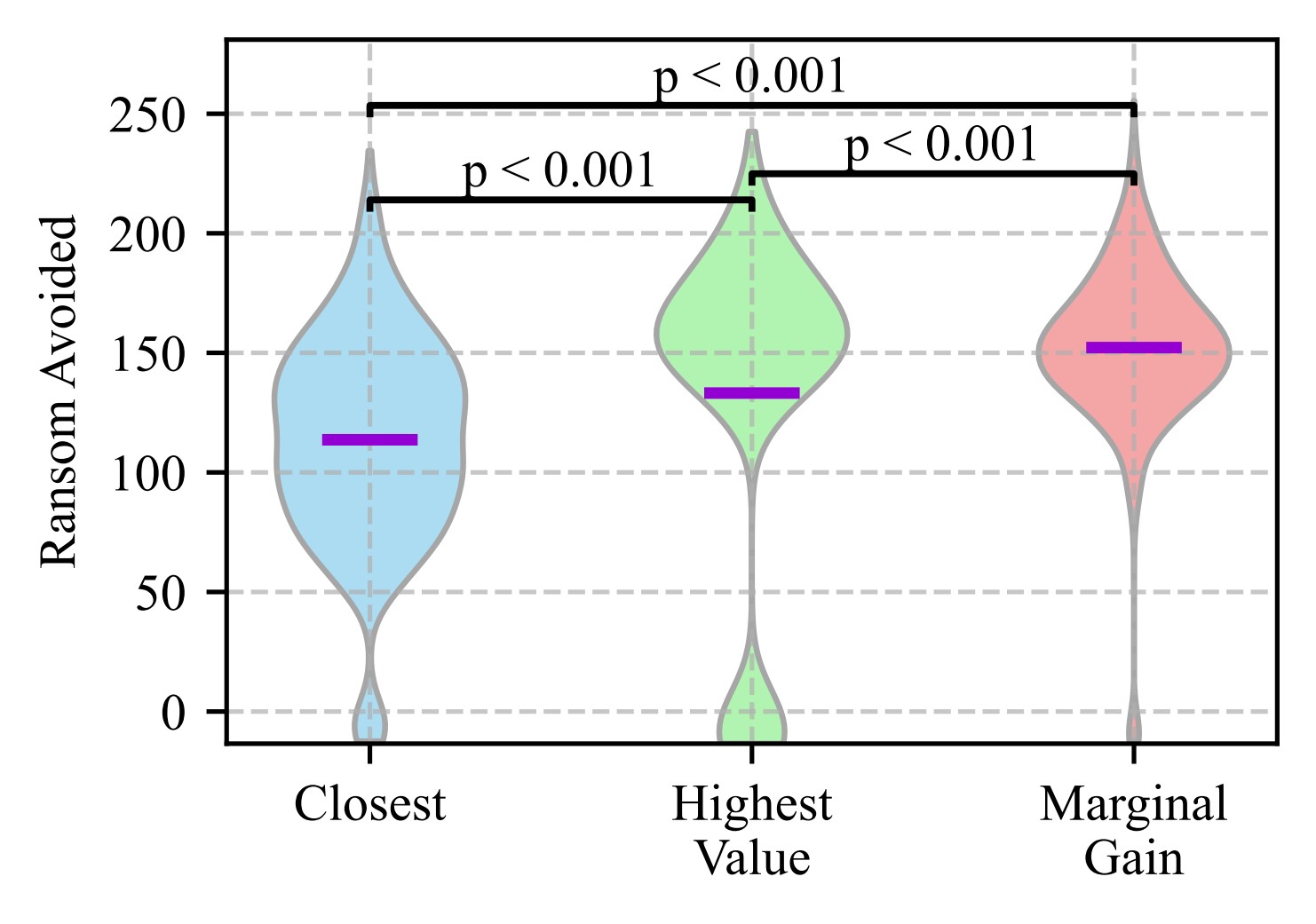}%
    \caption{Distributions of the ransom avoided by interdiction relative to baseline (no action) for different decision strategies (1000 trials; purple, horizontal lines identify means). The distributions are significantly different according to Kolmogorov-Smirnov tests.}
    \label{fig:simple-pirates}
\end{figure}

We consider  three strategies available to the agent: two heuristics (interdict closest attack; interdict attack on highest ransom ship) and a utilitarian assessment based on the expected marginal gain of interdiction. 
The agent can estimate the outcomes for each strategy and use these outcomes as a basis for choosing one attack to interdict. When its understanding is complete and accurate, the agent's assessments provide apt choices. In Figure~\ref{fig:simple-pirates}, which plots outcome distributions for the three strategies over 1000 scenario instances, the utilitarian strategy, which  has more complete knowledge of the situation (probability of successful interdiction and merchant ransom value combined, or ``marginal gain''), is better than the others (is selected in Step~4 of Algorithm~\ref{alg:bare-process-model}).

This scenario identifies knowledge relevant to enabling OAMNCC. The agent considers all its decision strategies because they each enable the agent to differentiate among candidate courses of action (Step~4). Further, marginal utility requires the knowledge to ``reframe'' duty constraints (orders) as utilitarian constraints (Rq: \textit{constraints \& frames}). %

\subsubsection{Merchants with Water Cannons}
This scenario is a variation of the interdiction scenario. We assume new information is available to the agent. The ship has recently received a command memo that indicates water cannons, which have previously been limited to installation on only the largest tankers, are now being installed on all sizes of tankers and container ships as well. Water cannons enable merchant vessels to mount an autonomous, reasonably effective defense against typical pirate attacks. 

\begin{figure}[t]
    \centering
    \includegraphics[width=0.991\columnwidth]{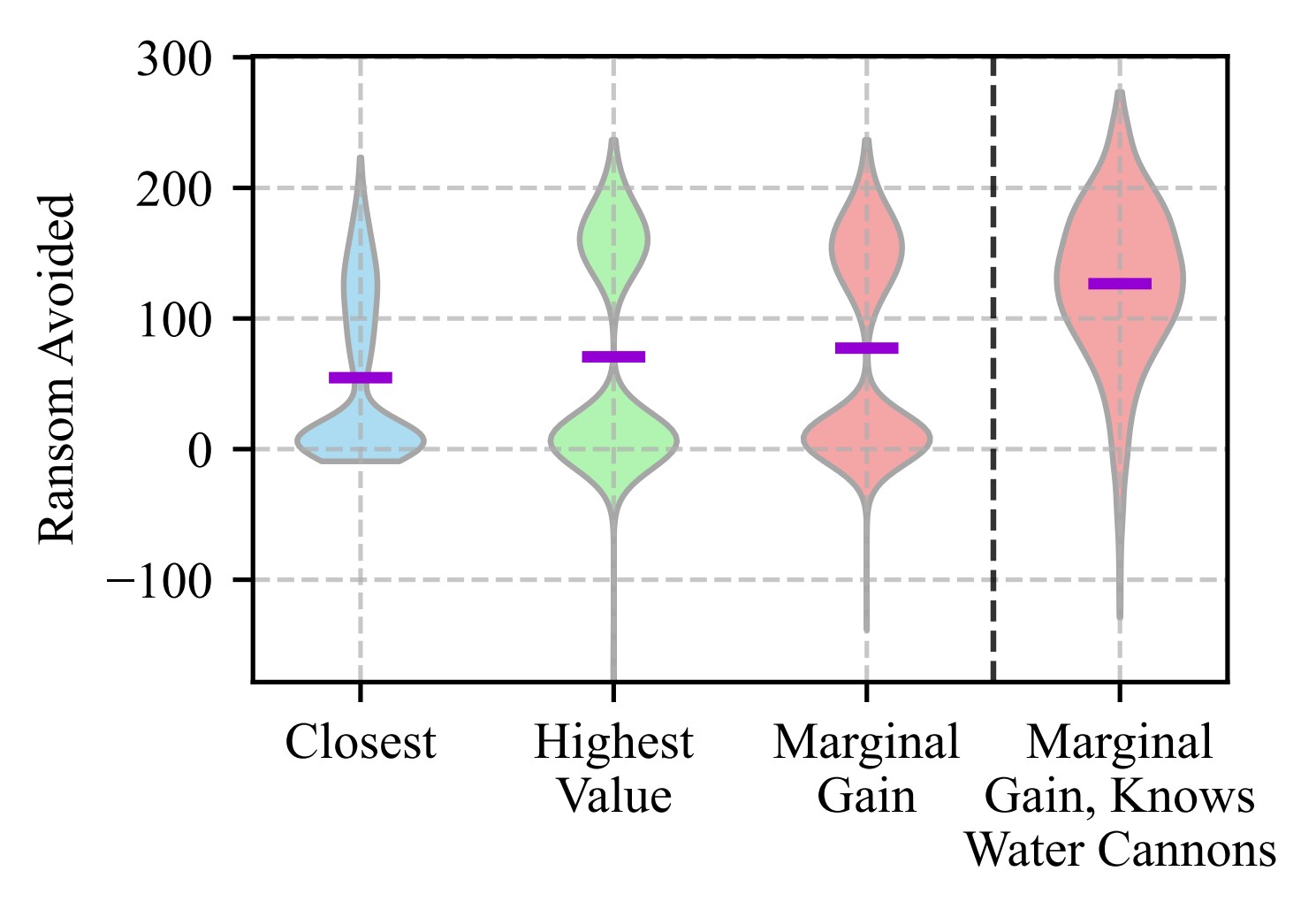}%
    \caption{Distributions of ransom avoided for each strategy when two merchant vessels can deploy water cannons. Left of dashed vertical line: the agent ignores water cannon capabilities.}
    \label{fig:water-cannons}
\end{figure}

What impact does this information have on the decision about which pirate to interdict?  First, let us assume that the agent is incapable of dynamically updating its statistical expectations to take into account the command memo. Under this assumption, the agent's assessment of the interdiction choice remains the same as in Figure~\ref{fig:simple-pirates}. However, in actuality, using the strategies defined previously, but ignoring the water-cannon memo, produces the distribution of outcomes in Figure~\ref{fig:water-cannons} (left of the dashed line). Failing to include water cannons in the utilitarian frame causes this gap. 

Now, assume the agent is capable of dynamically updating its expectations (Algorithm~\ref{alg:bare-process-model} Step~3). While it lacks the statistical confidence gained from policy training in this instance, the provenance of the memo, a command directive, leads it to modify its framing. The agent's expectations for this situation are illustrated to the right of the dashed line in Figure~\ref{fig:water-cannons}. Juxtaposing the differences in these outcomes demonstrates the impact of not taking the new information into account. Ships that could have defended themselves are protected, leaving others unnecessarily vulnerable.

This variation reemphasizes the continual need to update situation model knowledge (Rq: \textit{dynamic situation model}). However, it also suggests the need to assess information quality based on multiple factors. If the information about water cannons came from a random social media post, rather than command, should it have the same impact on the world model? Here, the provenance of the information is a key factor in whether/when to update one's world model. Assessment of \textit{information quality}, extending to \textit{provenance}, is necessary and emphasizes the critical role for meta-knowledge (and its online evaluation) in OAMNCC (Rq: \textit{information quality}). %
\begin{table*}[t!]
    \centering
    {\fontsize{9pt}{10.8pt}\selectfont
    \begin{tabularx}{.975\textwidth}{p{0.21\textwidth} X} 
        \toprule
        \textbf{Knowledge Type} & \textbf{Definition} \\
        \midrule

        \textbf{World Knowledge} & Knowledge of embodiment and constraints needed for human-aligned responses \\
        Constraints \& Frames (\textbf{CF}) & Knowledge of constraints, constraint frames, knowledge of how to apply  frame(s) to constraint(s) not already associated with that frame \\
        Expressive \mbox{Preferences} (\textbf{EP}) & Sensitivity of preferences to specific states of grounding of a constraint. Ability to prioritize among constraints directly and by constraint properties such as types (e.g., laws, etiquette) and sources (e.g., inferred from observation, written regulation) \\ %
        Action Affordances (\textbf{AA}) & Task-general specifications of affordances, recognition of applicability of affordance not explicitly associated with current  context \\
        Dynamic \mbox{Situation Model (\textbf{SM})} & Sensitivity to environment or memory enabling updated representations of  situation and near-term expectations (e.g., action models) \\
        \midrule

        \textbf{Metaknowledge} & Refers to other knowledge, knowledge that determines what information to trust and use \\
        Information Quality (\textbf{IQ}) & Reliability of other sources of knowledge through properties such as provenance or statistical certainty. May require the ability to act  to improve knowledge of quality \\
        Mitigation Utility (\textbf{MU}) & Whether the referent information or strategy will differentiate conflicted courses of action \\
        Conflict Structure (\textbf{CS}) & Type of conflict (i.e., from Table~\ref{tab:conflict_taxonomy}, inter-constraint or inter-constraint conflict, etc.)\\
        
        \bottomrule
    \end{tabularx}}
    \caption{A Taxonomy of Required Knowledge Types for Online Aligned Mitigation of Novel Constraint Conflicts}
    \label{tab:knowledge-types}
\end{table*}

\subsubsection{Piracy and vessel adrift}
In this scenario, there is an imminent threat of piracy on one merchant vessel, and there is simultaneously an unidentified mass spotted adrift. Here, we describe how integrating additional sources of knowledge contributes to improved outcomes but omit empirical results.

The drifting mass requires ship attention because adrift objects ordinarily require ship inspection to ensure safety in the shipping lanes and identify lost cargoes. Again, the agent faces a conflict due to temporal resource contention. If the ship does not immediately track the object, it may be difficult to find again. However, tracking and inspection would preclude timely interdiction of the pirates. Further, without some initial inspection, the actual value/cost of ignoring the mass is wholly unknown to the agent. 

In this case, consider the result if the agent assesses the applicability of a novel affordance (as part of evaluating relevant information in Algorithm\ref{alg:bare-process-model} Step~3). We adopt this idea from the original NPS scenarios, in which a new aerial drone capability is used to track a lifeboat adrift until direct assistance can be offered. Here, the agent tasks a ship-launched, aerial drone to fly to the mass and mark it with a beacon. This allows immediate interdiction and the likely ability to inspect the flotsam subsequently. Thus, considering alternative/new affordances enables resolving the temporal resource conflict. 

The resolution relies on knowledge of \textit{action affordances} outside of their usual context (lifeboat rescue) and of \textit{conflict structure} (temporal resource contention) (Rq: \textit{action affordances, conflict structure}). Further,  the course of action is being chosen based on knowledge of the conflict \textit{mitigation utility} of that action derived from the relation between the affordances of the action and the conflict type (Rq: \textit{mitigation utility}).

\section{Knowledge Requirements for OAMNCC}

These scenarios bring to light more detailed  knowledge requirements for OAMNCC, summarized in Table~\ref{tab:knowledge-types}.\footnote{The linked technical appendix extends the presentation of each item in the table and identifies prior, related work where that type of knowledge was researched.}
That is, the analysis from the prior section demonstrates that this knowledge is necessary for resolving conflicts across the space identified in Table~\ref{tab:conflict_taxonomy}. Although specific knowledge contents will vary across domains, we claim that these same kinds of processes and categories of knowledge and processes will generally be needed across domains. While we do not claim this list of requirements is sufficient for every conflict, these requirements further clarify the challenging scope for OAMNCC.

The more detailed analysis also allows us to map each type of knowledge to the process model introduced previously. 
Algorithm~\ref{alg:knowledge-laden-process-model} highlights where different types of knowledge are needed to influence the conflict mitigation process, which we discuss in more detail as follows:

Step 2: Assessing the conflict provides a basis for subsequently evaluating the relevance of other situation information. Assessment requires not only knowledge of conflict type but also diagnostic knowledge, such as identifying if a conflict is \textit{intra-constraint} (e.g., conflicting instances of the same duty). Significantly, this step  determines what situational information to take into account when the features in existing policies may be insufficient to resolve a conflict.%

Step 3: Generating candidate, off-policy courses of action requires the agent to identify and to integrate novel sources of knowledge relevant to the situation. This  includes extending the state representation to incorporate features now determined to be salient (\textbf{SM}) and reasoning about action affordances (\textbf{AA}). This step may also require reinterpretation of constraints (\textbf{CF}) outside of their typical encoding. Further, any relevant information used as a decision basis should be evaluated for quality, potentially using situationally available assessment of that quality if apt (\textbf{IQ}).

\begin{algorithm}[tb!]
\caption{The different knowledge that OAMNCC processing steps demand (ref: Table~\ref{tab:knowledge-types} for bold labels)}
\label{alg:knowledge-laden-process-model}
\begin{algorithmic}[1] %
\STATE Recognize novelty of the conflict \\ (similar to ``out of distribution'' assessment)
\STATE Assess conflict type and structure (\textbf{CS})  %
\STATE Evaluate what situational information might be relevant to the conflict (\textbf{IQ; CF, EP, AA, SM})
\STATE Propose candidate conflict-mitigating courses of action, evaluate them, and select one (\textbf{MU, EP})
\STATE Execute the selected course of action, monitoring (\textbf{CS}) for resolution of conflict
\end{algorithmic}
\end{algorithm}

Steps 4 \& 5: Candidates generated via information identified as relevant from Step~3 are then evaluated and one course of action chosen. Evaluation gauges the likelihood that the courses of action will resolve the triggering conflict(s) or extend the agent's situational understanding, such as through active sensing (\textbf{MU}). This capability includes the ability to determine if a candidate will not contribute to mitigation. For example,  constraint prioritization is irrelevant to \textit{intra-constraint} conflicts. Additionally, the process may iterate between evaluating available information (Step~3) and candidates (Step~4), especially when candidates are judged insufficient. Ultimately, candidate selection (Step~4) must justify its selection using preferences sufficiently expressive to encode human preference expressivity (\textbf{EP}). Finally, detection that a conflict is successfully resolved depends on the capability to monitor actual outcomes (\textbf{CS}).

Implementation challenges are now more evident. First, some requirements are in tension: the need for online response and the potentially expensive evaluation of situational information, expressive preferences and comprehensive coverage over all conflict types, the use of information based on its quality or the role it can play in mitigation. Second, these tensions cannot just be resolved once, but must be frequently re-evaluated. Different contexts will offer differing accessibility to situation knowledge and metaknowledge. Further, dynamics may include potentially adversarial responses to the agent's courses of action (e.g., pirates may begin to use ruses and feints in response to the ship's strategy). Third, underlying these mitigation challenges is the vast scope of acquiring, representing, and reliably retrieving the necessary knowledge, regardless of the specific implementation-level algorithms. Nonetheless, the combination of these requirements more fully characterizes the solution space for comprehensive conflict mitigation that aligns with human expectations and values.

\section{Related Work}

The knowledge requirements from the previous section can also be used to organize state-of-the-art approaches and to assess which methods satisfy which requirements. In this section, we review and organize past work, using the requirements as a guide. Despite a diversity of approaches, a common theme in  related work is evident: they generally fail to integrate explicit metaknowledge when encountering a novel situation. 

Many examples of past work address various types of conflicts directly, but incompletely with respect to the full set of requirements we identify. For example, some address a subset of all possible conflicts \cite{banerjee_dynamic_2024,choi_resolving_2025,briggs_sorry_2015,lu_ethically_2024}, some rely on pretrained response selection \cite{awad_when_2024,simpkins_composable_2019}, and some are limited in both ways \cite{censi_liability_2019,loreggia_making_2022}. As one example, SEP-nets \cite{awad_when_2024} excel as representations explicitly designed to capture human preference expressivity, conditioned on a specific context. These methods vary in precise capabilities and will often be sufficient in closed domains. Table~\ref{tab:related_work_summary} (and the Appendix) contain further characterization of the specific types of knowledge they support (and do not).

{\setlength{\tabcolsep}{1mm}
\begin{table*}[t]%
    \centering
    \begin{minipage}{0.95\textwidth} 
    \centering
    \renewcommand{\theadfont}{\bfseries} %
    {\fontsize{9pt}{10.8pt}\selectfont
    \begin{tabular}{@{}lccccccc@{}} 
        \toprule
        \thead[l]{Related Work} & 
        \thead{Constraints \\ \& Frames} & 
        \thead{Expressive \\ Preferences} & 
        \thead{Action \\ Affordances} & 
        \thead{Situation \\ Model} & 
        \thead{Conflict \\ Structure} & 
        \thead{Information \\ Quality} & 
        \thead{Mitigation \\ Utility} \\
        \midrule
        
        \citet{censi_liability_2019} & $\times$ & $\sim$ & $\times$ & $\times$ & $\times$ & $\times$ &  $\times$ \\
        \citet{briggs_sorry_2015} & $\times$ & $\sim$ & $\sim$ & $\sim$ & $\times$ & $\times$ &  $\sim$ \\
        \citet{awad_when_2024} & \checkmark & \checkmark & $\times$ & $\sim$ & $\times$ & $\times$ &  $\times$ \\
        \citet{lu_ethically_2024} & $\times$ & $\sim$ & $\times$ & $\sim$ & $\times$ & $\sim$ & $\sim$ \\
        \citet{loreggia_making_2022} & $\sim$ & $\sim$ & $\times$ & $\sim$ & $\times$ & $\times$ & $\times$ \\
        \citet{simpkins_composable_2019} & $\sim$ & \checkmark & $\sim$ & $\sim$ & $\sim$ & $\times$ & $\times$ \\
        \citet{choi_resolving_2025} & $\times$ & $\times$ & $\sim$ & $\times$ & $\times$ & $\times$ & $\sim$ \\
        \citet{banerjee_dynamic_2024} & $\times$ & $\times$ & \checkmark & $\sim$ & $\times$ & $\times$ & $\sim$ \\
        \citet{mclaren_computational_2006} & \checkmark & \checkmark & $\sim$ & $\times$ & $\sim$ & $\times$ & $\sim$ \\
        \citet{aha_goal_2018} & $\sim$ & $\sim$ & $\sim$ & \checkmark & $\times$ & \checkmark & $\sim$ \\
        
        \bottomrule
    \end{tabular}
    }
    \end{minipage}
    \caption{Summarizing related work in terms of coverage of the knowledge requirements. A \checkmark indicates satisfaction of a knowledge requirement. An $\times$ indicates lack of this type of knowledge. A $\sim$ indicates partial satisfaction of the requirement.}
    \label{tab:related_work_summary}
\end{table*}}

Generally, there is increasing recognition of the importance of, for deployed, long-lived agents, online novelty detection, including, but broader than, detecting out-of-distribution situations \cite[e.g.,][]{haider_can_2024,mohan_domain-independent_2024,kejriwal_challenges_2024,goel_novelgym_2024}. 

We turn now to methods that detect and respond to out-of-distribution or novel situations. These methods can also likely apply novelty detection toward mitigating different types of conflicts, although OAMNCC per se is not the focus.
For example, \citet{mclaren_computational_2006}  applies case-based reasoning to ethical dilemmas. Their method reuses the reasoning process and decision rationale from a similar, past dilemma in order to respond to a novel one. This mechanism can be sensitive to action affordances, and \emph{depending on the case-base and query mechanism}, can reflect complex preferences and use of different constraint and frame knowledge in a novel situation. However, the approach depends on static, pre-defined, similarity-based case retrieval. This retrieval may be insensitive to subtly different contexts, limiting the use of meta-knowledge (especially mitigation utility) as the basis for selection among possible responses to a conflict in a situation. Further, given the reliance on similarity-based retrieval, it is unclear if action affordances can be transferred from substantially different cases. %

Goal reasoning agents \cite{aha_goal_2018} can autonomously manage their own goals according to a process conceptually similar to Algorithm~\ref{alg:bare-process-model}. Goals can have varied semantics analogous to multiple constraint frames, and can trigger information gathering. The method potentially enables multiple goal selection strategies, including ones suitable for mutual exclusivity conflicts \cite{smith_choosing_2004}. Related, goal-reasoning research describes autonomous goal selection in the context of novelty with ``motivators'', but leaves trustworthy online motivators to future work \cite{johnson_goal_2018_different}. While spanning many requirements, these strategies are not chosen according to situational information about the type of conflict, limiting any use of conflict structure knowledge, and thus also relevance and mitigation utility knowledge in turn. Additionally, we were unable to identify examples in which a goal reasoning system attempts to reframe its interpretation of constraints. Overall, because this research is dispersed among a few (sometimes domain-specific) methods, it is unclear how to integrate them to specifically support the required knowledge altogether.

In summary, no prior work fully supports all knowledge requirements. However, a combination of case-based and goal reasoning approaches \cite[e.g.,][but without assumption of human interaction]{floyd_trust-guided_2015}, with more flexible processing for their individual components, may approach the necessary knowledge support (also integrated with additional goal reasoning strategies related to conflict structure).
Generally, a goal reasoning approach appears to be a potentially apt starting point for implementation, given how much of the problem scope it addresses.

\section{Conclusion}
AI as a field is seeking to enable long-lived, autonomous agents that act in accordance with human expectations and values. To achieve this goal, agents must successfully and adaptively comply with many kinds of constraints (normative and otherwise), and, when they inevitably arise during deployment, resolve conflicts between and among those constraints, also in ways aligned with human expectations. This paper presented a knowledge-level analysis of what such an aligned conflict mitigation strategy entails, both in terms of types of conflicts that may arise and the types of knowledge that an agent requires to navigate these conflicts. The analysis demonstrates the significant challenge aligned conflict mitigation presents, where an agent will have to contend with incomplete policy knowledge and underspecified decisions, while remaining sensitive to the diverse and dynamic situation in which it is embedded.

A perspective that may enable progress on this challenge is to view a novel, unanticipated conflict as an instance of an \textit{ill-structured problem} \cite{newell_heuristic_1969,simon_structure_1973}, a problem in which it is unclear what information and actions are relevant. An agent may initially lack the capacity to formulate (let alone solve) an ill-structured problem. These problems often lead to a reconceptualization of the situation. For an agent to be able to safely and robustly act in an open, unpredictable world in ways aligned with human values, this perspective, drawing from the analysis presented here, suggests the need for dynamic metacognition, or the ability to reason about and adapt an agent's own problem-solving framework. %

Thus, in terms of the process models sketched here, in many cases a first step should not be an attempt to solve the novel conflict (a kind of ill-structured problem), but instead to determine what problem space the agent should use to attempt to address the conflict. We expect reconceptualization will need to  utilize metacognitive reasoning to search for an appropriate problem space, which may often include reconsideration and reframing of the original problem. The example of reframing a duty as a cost function (and vice versa), as we presented in the scenarios, is one example of such reframing. We thus hypothesize that  dynamic runtime metacognition, capable of responding to specific situational information, will be a key feature of solutions to online, aligned mitigation of novel constraint conflicts.

\section{Acknowledgments} %

This work was supported by the Office of Naval Research, contract N00014-22-1-2358. The views and conclusions contained in this document are those of the authors and should not be interpreted as representing the official policies, either expressed or implied, of the Department of Defense or Office of Naval Research. The U.S. Government is authorized to reproduce and distribute reprints for Government purposes notwithstanding any copyright notation hereon. We acknowledge and thank the anonymous reviewers for their constructive suggestions.

\bibliography{abc-references,aaai26}

\include{appendix}

\end{document}

%% file: appendix.tex
\onecolumn
\appendix
\newcommand{\apptitle}[1]{\noindent{\centering\LARGE #1}}
\apptitle{\hspace{.38\linewidth} Technical Appendix}

\renewcommand{\thefigure}{A-\arabic{figure}}
\setcounter{figure}{0}
\renewcommand{\thetable}{A-\arabic{table}}
\setcounter{table}{0}

\section{Scenarios Illustrating Requirements for Knowledge}
The following sections detail the scenarios outlined in the body of the paper. Each scenario includes the following elements:
\begin{itemize}
    \item Description: Narrative description of the scenario, including a figure for some scenarios.
    \item Conflict Type: Short description of the type of constraint conflict the scenario is designed to elicit in agent reasoning. Conflict types correspond to specific types of conflicts listed in Table~\ref{tab:conflict_taxonomy}.
    \item Relevant Information: The external source of information, knowledge, expectation that (we assume) is salient to human expectation (and is not already integrated into the agent's existing policy knowledge).
    \item Outcome Discrepancy: A short description of what happens when the agent omits, lacks access, or otherwise fails to take into account knowledge/information that is relevant to making choice(s) aligned with human expectations.
    \item Required Knowledge:  A description/enumeration of the specific knowledge (and the type of knowledge) that the agent requires access to in order to accomplish aligned and effective outcomes in the scenario and circumvent the outcome discrepancy. The types of knowledge correspond to the knowledge requirements listed in Table~\ref{tab:knowledge-types}.
\end{itemize}

For scenarios where empirical data were collected, we developed a simple, fixed-increment time advance simulation represented various kinds of sea vessels. The naval drone ship was inspired by the US Navy's large autonomous surface ship (LUSS) program. However, specific parameters in the sailor overboard scenario for general capabilities, speed, and fuel consumption were adopted from public data on the Arleigh Burke class of US Navy destroyers. While LUSS vessels are significantly smaller than Arleigh Burke class ships, LUSS performance data is not readily accessible to the public and the destroyer is a crewed vessel with similar missions to those described here. 

Events such as the pirate attacks and sailor rescue are modeled with simple probability distributions reflecting estimates from available data. For example, for the pirate scenario, we used published estimates of pirate attack success rates and the typical length of engagements \cite{leontarakis_piracy_2015}. For interdiction, the simulation assumes the pirate attack fails if the drone ship projects force at the location of the merchant vessel and the pirates have not succeeded in their attack.

The simulation was written in standard Python. Specific scenarios were created to support the different uses cases described in the paper. For example, for the pirate interdiction scenario, we generated 1000 random instances of the scenario, where the merchants were present in the sea lane but sufficiently separated from one another that the destroyer at full speed could not reach both ships in 30m (the length of time of pirate attacks as above).  Data were collected for each scenario instance and then aggregate statistics were computed to create the distribution plots in the paper. 

Simulation source code, data analysis scripts, and data are available at: \url{https://github.com/Center-for-Integrated-Cognition/OAMNCC}.

\subsection{Sailor Overboard Scenario}
\subsubsection{Description}
In this scenario, we assume there is an urgent/high priority need to return to base (RTB). During transit back to base, someone is lost at sea (sailor overboard). Sometime later, the agent discovers a sailor was lost overboard.

The scenario is designed so that the overboard alert occurs 20nm from the ship's destination. The agent will then backtrack and search until its fuel runs so low that reaching base is endangered. By design, the sailor overboard occurs between 5 and 15 nm behind the the alert location. Ignoring noise in fuel consumption for different activities, the scenario is designed so that the ship can always safely backtrack 10nm and still reach port before its fuel is exhausted. Thus, generally, the agent can only recover the 50\% of the time and still return to base.

The simulation uses different rates of fuel consumption for transit, backtracking, and search patterns and models noise/variability in the rates.

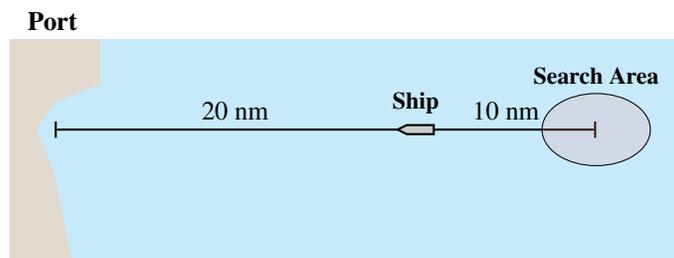
\begin{figure}[hb]
    \centering
\begin{tikzpicture}[scale=0.12]
\fill[cyan!20] (-5,5) rectangle (70,30);
  \fill[brown!50!gray!30] (-5,5) -- (-5,30) -- (5,30) -- (5,25) -- (0,23) -- (-2,20) -- (0,15) -- (1,10) -- (2,5) -- cycle;
  \node[align=center, anchor=south west] at (-4,30) {\textbf{Port}};

  \coordinate (port) at (0,20);
  \coordinate (ship) at (40,20);
  \coordinate (searchCenter) at (60,20);

  \draw[|.-., thick] (port) -- (ship.west) node[midway, above, sloped] {20 nm};
  \draw[.-.|, thick] (ship) -- (searchCenter) node[midway, above, sloped] {10 nm};

  \draw[fill=gray!40, thick] 
    (38,20) -- (39,20.5) -- (42,20.5) -- (42,19.5) -- (39,19.5)--cycle;
  \node at (40,23) {\footnotesize \textbf{Ship}};

  \fill[red!40, opacity=0.2] (60,20) ellipse (6 and 4);
  \draw (60,20) ellipse (6 and 4);
  \node at (60,26) {\footnotesize\textbf{Search Area}};
\end{tikzpicture}
    \caption{Schematic Representation of the Sailor Overboard Scenario.}
    \label{fig-app:sailor-overboard}
\end{figure}

Figure~\ref{fig-app:sailor-overboard} illustrates the basic geometry of the scenario. The sailor location is randomly located within the search area, meaning that the ship can generally only reach the midpoint of the search area before its fuel reaches the minimum required to safely return to port.

\subsubsection{Conflict Type(s)}
\begin{itemize}
    \item Conflict between a goal/order (RTB) and a duty (``leave no sailor behind.''). Equivalently, these can be cast as conflicting constraints.
    \item Mutual Exclusivity (resource contention) and uncertainty further complicate the decision calculus but are not the main source of conflict.
\end{itemize}

\subsubsection{Relevant Information}
\begin{itemize}
    \item Awareness/understanding of the conflict in duties
    \item Awareness/understanding that the relative importance of the duties are not known/readily discoverable
    \item Whether or not the sailor has been spotted and rescue operations have commenced at the time bingo fuel is reached
\end{itemize}

\subsubsection{Outcome Discrepancy}
Undesirable outcomes:
\begin{itemize}
    \item Searching no matter what, until RTB is impossible, even when the sailor has not been spotted.
    \item Abandoning rescue attempts once the sailor is spotted (The scenario is designed so that, once spotted, the sailor can always, eventually, be safely rescued with the probability of rescue success increasing each time step.)    
\end{itemize}

\subsubsection{Implied useful/required knowledge}
This refines the ``expressive preferences'' requirement to specifically include preferences sensitive to particular grounding statuses of a constraint.

\begin{figure}
    \centering
    \includegraphics{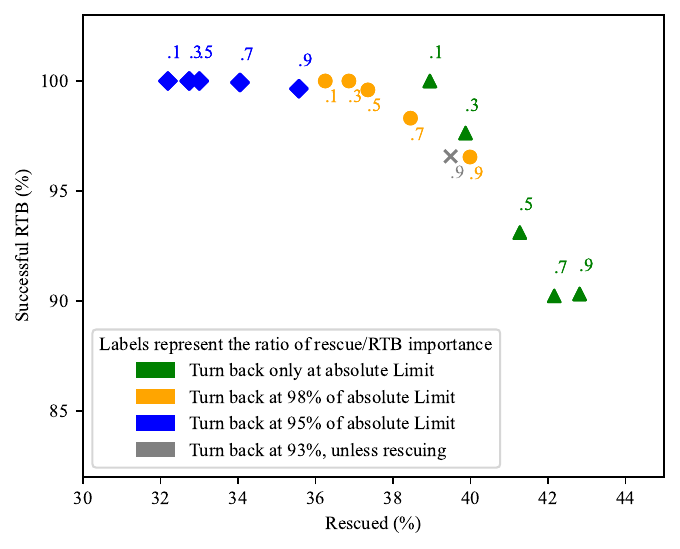}
    \caption{Utilitarian assessment of sailor overboard scenario. Colored dots represent different safety margins. Labels over marks provide the ratio of rescue/RTB importance. $\times$ represents a policy that never leaves behind a sailor once spotted.}
    \label{fig:placeholder}
\end{figure}

\subsection{Piracy Interdiction Scenarios}
\subsubsection{Description} In the remaining demonstrations, we envision a scenario where pirate attacks occur on a shipping lane, with only one local naval ship available to interdict the pirates, and four merchant ships in that lane are under imminent threat. Given the situation design, the captain's ship will only be capable of attempting to interdict one attacking pirate band. (We assume by the time the single interdiction is complete, the other three pirates attack will have resolved themselves one way or the other for the remaining ships).

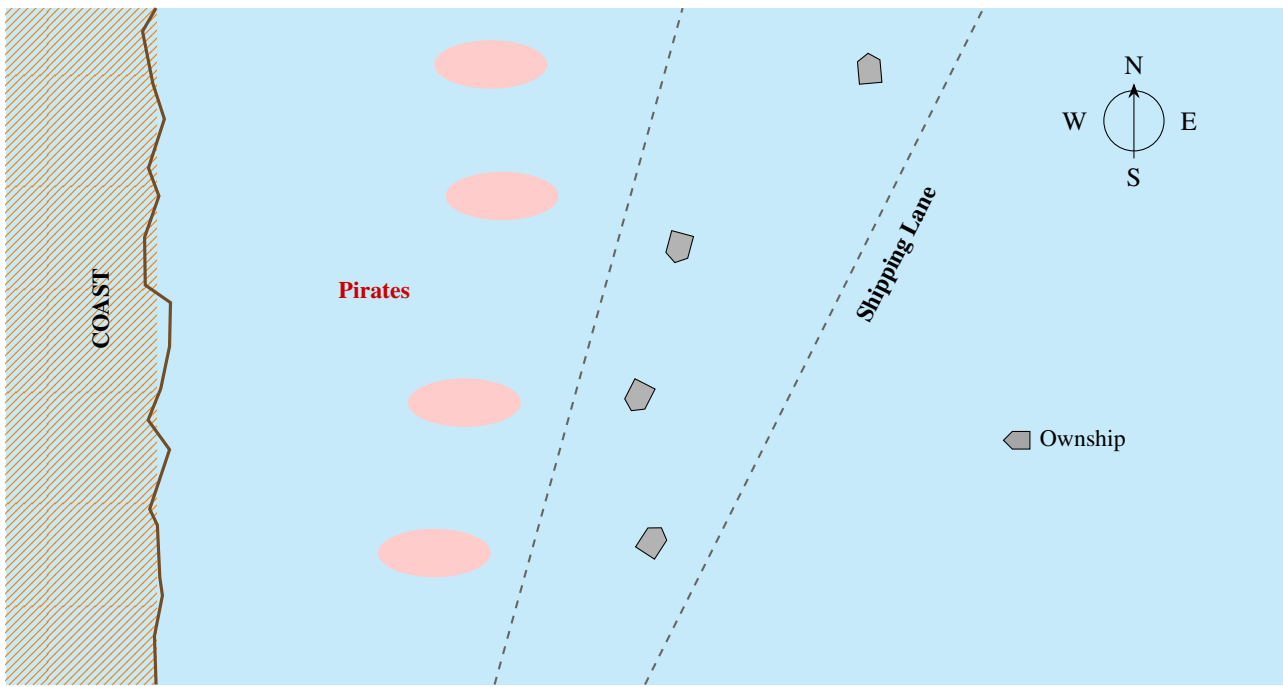
\begin{figure}[thb!]
    \centering
\begin{tikzpicture}[
    ship/.style={
        draw,
        fill=gray!70,
        signal,
        signal to=west,
        minimum height=.2cm%
    },
    pirate/.style={
        ellipse,
        fill=red!20,
        minimum width=1.5cm,
        minimum height=.5cm,
        inner sep=6.5pt
    }
]

\fill[cyan!20] (-1,0) rectangle (16,9);

\fill[pattern=north east lines, pattern color=orange!80!brown] (-1,0) rectangle (1,9);
\draw[decoration={random steps, segment length=5mm, amplitude=2mm}, decorate, very thick, brown!60!black]
    (1,0) -- (1,9);
\node[rotate=90, anchor=south, font=\small] at (0.5, 5) {\textbf{COAST}};

\draw[dashed, thick, black!60] (5.5,0) -- (8,9);
\draw[dashed, thick, black!60] (7.5,0) -- (12,9);
\node[rotate=63, anchor=south, font=\small] at (11.1, 5.6) {\textbf{Shipping Lane}};

\node[pirate] at (4.7, 1.75) {};
\node[pirate] at (5.1, 3.75) {};
\node[pirate] at (5.6, 6.5) {};
\node[pirate] at (5.45, 8.25) {};
\node[red!80!black, font=\bfseries\small, anchor=east] at (4.5, 5.25) {Pirates};

\node[ship, label={[font=\small]right:Ownship}] at (12.5, 3.25) {};

\newcommand{\merchantship}[3]{
  \begin{scope}[shift={(#1,#2)}, rotate=#3]
    \draw[fill=gray!60]
      (0,.4) --          %
      (.15,.3) --
      (.15,0) -- %
      (-.15,0) -- %
      (-.15,.3) --
      cycle;
  \end{scope}
}

\merchantship{7.5}{1.75}{-33}   %
\merchantship{7.5}{4}{153}  %
\merchantship{8}{6}{165}  %
\merchantship{10.5}{8}{5}

\begin{scope}[shift={(14,7.5)}]
    \draw[-{Stealth[length=2mm]}] (0,-0.5) -- (0,0.5) node[above] {N};
    \draw (0,-0.5) node[below] {S};
    \draw (-0.5,0) node[left] {W};
    \draw (0.5,0) node[right] {E};
    \path[draw] (0,0) circle (0.4cm);
\end{scope}
\end{tikzpicture}
\caption{Schematic Representation of the Piracy Interdiction Scenarios.}
\end{figure}

\subsubsection{Conflict Type(s)}
\begin{itemize}
\item{Conflict between four instances of the same standing order (duty).}
\item{Mutual Exclusivity (temporal resource contention), and uncertainty further complicates the decision calculus but is not the main source of conflict.}
\end{itemize}

\subsubsection{Relevant Information}
\begin{itemize}
    \item{Conflict between four instances of the same standing order (duty).}
    \item{Mutual Exclusivity (temporal resource contention), and uncertainty further complicates the decision calculus but is not the main source of conflict.}
\end{itemize}
situational information about distance, manifests of the ships, knowledge of how to reframe duties as utilitarian-like, knowledge that utility-framing can break the tie
\subsubsection{Outcome Discrepancy} Undesirable Outcomes:
\begin{itemize}
    \item{Uninformed random selection}
    \item{Doing nothing}
    \item{Using a subset of available knowledge}
\end{itemize}
\subsubsection{Implied useful/required knowledge}
This refines the constraint and frame knowledge to include an ability to reframe a constraint with another frame. It also shows that it's useful to select a mitigation response based on how it connects specifically to the conflict structure (i.e. tied instances of the same duty are readily resolved if given a utility framing). This example itself also shows a diagnostic aspect of conflict structure that is not covered by the taxonomy: intra-constraint conflicts in general preclude the use of cross-constraint prioritization.

\begin{figure}[htb!]
    \centering
    \includegraphics{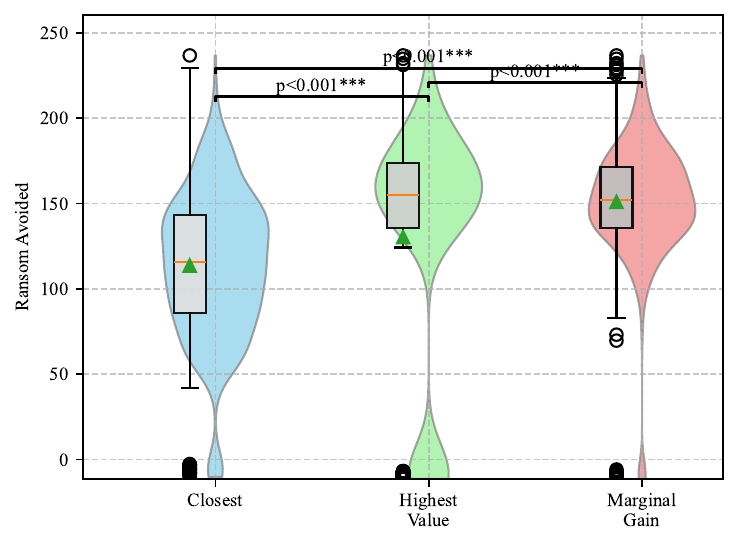}
    \caption{Distributions of the ransom avoided by interdiction relative to baseline (no action) for different decision stategies (1000 trials; green triangles identify means, and each distribution is significantly different according to a Kolmogorov-Smirnov test.).}
    \label{fig:simple-pirates-big}
\end{figure}

\subsection{Merchants with Water Cannons}
\subsubsection{Description} The situation is the same as before, but some merchant ships are equipped with defensive capabilities.
\subsubsection{Conflict Type(s)} (Same as before:)
\begin{itemize}
    \item{Conflict between four instances of the same standing order (duty).}
    \item{Mutual Exclusivity (temporal resource contention), and uncertainty further complicates the decision calculus but is not the main source of conflict.}
\end{itemize}
\subsubsection{Relevant Information}
\begin{itemize}
    \item{Defensive water cannon capabilities on different ships than before}
    \item{Information is derived from a memo from command}
\end{itemize}
\subsubsection{Outcome Discrepancy} Undesirable outcome:
\begin{itemize}
    \item{Choosing to defend a ship that can defend itself.}
\end{itemize}
\subsubsection{Implied useful/required knowledge}
This establishes a need for dynamic situation model knowledge that can take into account situational sources of knowledge, and evaluation of information quality that extends to provenance.

\subsection{Piracy and vessel adrift}
\subsubsection{Description}
There is an imminent threat of piracy on a merchant vessel, and there is unidentified flotsam spotted adrift. There is generally a duty to inspect the unidentified object, and a higher priority or at least urgency duty to interdict piracy. 
\subsubsection{Conflict Type(s)}
\begin{itemize}
    \item{Conflict between four instances of the same standing order (duty).}
    \item{Mutual Exclusivity (temporal resource contention), and uncertainty, as the importance of one duty is unknown.}
\end{itemize}
\subsubsection{Relevant Information}
\begin{itemize}
    \item{Action affordance knowledge: A drone can be used as a beacon to facilitate relocation of the adrift mass.}
\end{itemize}
\subsubsection{Outcome Discrepancy}
\begin{itemize}
    \item{Undesirable: Foregoing interdiction to inspect the mass to achieve certainty.}
    \item{Acceptable, but not ideal: interdicting piracy according to preexisting general prioritization}
\end{itemize}
\subsubsection{Implied useful/required knowledge}
This reaffirms a need for knowledge of conflict utility to include expectations that a resolution strategy will address the structure of the specific conflict. Additionally, knowledge specifically of an action's affordances need to be available outside of the action's usual context.

\section{Expanded Description of Knowledge Requirements}
We describe the knowledge requirements in a little more detail and clarify some guidelines for our evaluation of related work with respect to them.

\subsection*{Base Knowledge Requirements}
\begin{description}
    \item[Constraint \& Frame Knowledge:] The ability to handle multiple, computationally distinct constraint frames (e.g., utility-based vs. rule-based), as described in calls for hybrid ethical architectures \cite{kuipers_how_2018}. It extends to a capability to dynamically reframe a conflict, such as in the ``Piracy Interdiction'' scenarios. In evaluating related work, a partial check ($\sim$) is given for handling distinct frames; a full check (\checkmark) requires the ability to dynamically reframe a constraint.
    \item[Expressive Preferences:] The ability to represent highly expressive, context-sensitive preferences that can be conditioned on multiple features and specific real-time grounding states (as demonstrated in the ``Sailor Overboard'' dilemma). This also extends to ability to prioritize among constraints directly. SEP-nets capture this level of expressivity well.
    \item[Action Affordances (Policy Transfer):] The ability to achieve compositional flexibility by dynamically creating a novel plan or sequence of actions from knowledge of how affordances of available actions relate to a conflict situation. This could range from selectional flexibility in choosing among existing policies to compositional flexibility, as in the constructed use of an action in the ``Vessel Adrift'' scenario. Both planning and online policy transfer are apt.
    \item[Situation Model Knowledge:] The ability to use and dynamically update a rich, predictive model of the environment and near-term expectations based on available information. A partial check ($\sim$) is given for systems that can update their knowledge of the current situation based on observations (as in a POMDP), while a full check (\checkmark) is reserved for systems that can also update their model based on new, abstract world knowledge at runtime (as in the ``Water Cannon'' scenario).
\end{description}

\subsection*{Meta-Knowledge Requirements}
\begin{description}
    \item[Knowledge of Information Quality:] The agent must be sensitive to the quality of other sources of knowledge, through properties such as provenance, statistical certainty, or reliability. Additionally, an agent should be capable of using available actions (i.e. active sensing) to improve its knowledge of quality.
    \item[Knowledge of Conflict Structure:] As a key input to assessing mitigation utility and relevance of sources of base knowledge, this is the knowledge of the type of conflict (both in terms of the constraint conflict taxonomy and also structural features like inter/intra-constraint) used to characterize why compliance is challenged in the current situation. It is intended to support differential response to conflicts.
    \item[Knowledge of Conflict Mitigation Utility:] The agent must predicate the enaction of a mitigation response on expectation it will mitigate the conflict, especially assessing that one strategy will mitigate better or contribute improved knowledge of the conflict situation than another. A partial check ($\sim$) is given for systems with a single, triggered response; a full check (\checkmark) is given for systems that can choose between multiple distinct mitigation strategies.
\end{description}

\section{Detailed Justifications for Related Work Evaluation}
\label{app:justifications}

This following provides more detailed, snippet-style justifications for the ratings presented in the main paper in Table~\ref{tab:related_work_summary}.

\subsection{\citet{censi_liability_2019}} Rulebooks pre-encode relative priority among constraints, making them suitable for training-time mitigation of specifically unsatisfiability conflicts.
\subsubsection{Constraints \& Frames}
($\times$) While rules can refer to different forms of constraint (e.g., legal, ethical), this does not extend more broadly to more computationally distinct constraints about values and utility.
\subsubsection{Expressive Preferences}
($\sim$) A predefined hierarchy captures some prioritization knowledge, but lacks expressive context-sensitive nuance.
\subsubsection{Action Affordances}
($\times$) Follows a fixed set of rules. Won't compose actions to resolve a conflict.
\subsubsection{Situation Model}
($\times$) Not integrated with an ability to update state representation/expectations.
\subsubsection{Conflict Structure}
($\times$) Only sensitive to Unsatisfiability conflicts.
\subsubsection{Information Quality}
($\times$) Not integrated with the assessment of information.
\subsubsection{Mitigation Utility}
($\times$) The resolution is not assessed with respect to the conflict situation.

\subsection{\citet{briggs_sorry_2015}} This work explores how a robot can appropriately reject directives when the implied actions would violate normative permissibility, providing conflict mitigation related specifically to obligations and permissions in human interaction. This is not cast within the context of resolving conflicts (especially with unavoidable violation) that arise from autonomous agent activity.
\subsubsection{Constraints \& Frames}
($\times$) Constraints are limited to obligations and permissions.
\subsubsection{Expressive Preferences}
($\sim$) The preference to reject a command is context-sensitive, as it is conditioned on a simulation of the action's outcome from the current state. However, the preference structure itself is a simple, absolute rule
\subsubsection{Action Affordances}
($\sim$) Some knowledge of composition of action affordances is used in determining a response, but not to potentially create new mitigation responses.
\subsubsection{Situation Model}
($\sim$) Simulates the outcome of a human directive, but does not autonomously update its situation model.
\subsubsection{Conflict Structure}
($\times$) Focused only on mitigating one conflict type.
\subsubsection{Information Quality}
($\times$) Not described as integrated with an ability to assess information quality.
\subsubsection{Mitigation Utility}
($\sim$) The refusal to follow a command is triggered when appropriate, but there are no additional potential response strategies.

\subsection{\citet{awad_when_2024}} SEP-nets are representations explicitly designed to capture much of human preference expressivity relative to constraints, conditioned on specific context, and excel at this. However, this work does not present methods for learning mitigation for novel conflicts online, or selection among conflict responses sensitive to conflict structure.
\subsubsection{Constraints \& Frames}
(\checkmark) The high expressivity of SEP net representations can capture many constraints, and the conditional preferences can implicitly support behavior that acts as though constraints have been reframed.
\subsubsection{Expressive Preferences}
(\checkmark) SEP-nets are used to model complex scenario evaluations in determining how to respond.
\subsubsection{Action Affordances}
($\times$) Learns a policy for when to break a rule, but cannot flexibly use action affordance knowledge to determine a response.
\subsubsection{Situation Model}
($\sim$) The model is conditioned on various scenario features, but does not update its situation model at runtime.
\subsubsection{Conflict Structure}
($\times$) They present learning of a policy for a single conflict type and do not classify conflict situations.
\subsubsection{Information Quality}
($\times$) The learned model does not assess the quality of new situational information.
\subsubsection{Mitigation Utility}
($\times$) The model does not assess the utility of mitigation strategies.

\subsection{\citet{lu_ethically_2024}} This work directly considers the problem of supporting Aristotelian virtue ethics values of courage and prudence and how they conflict in the context of partial observability. It provides some support for mitigating mutually exclusivity and uncertainty conflicts in a predetermined way, and is currently limited to support for one ethical frame.
\subsubsection{Constraints \& Frames}
($\times$) This work is focused on a single type of constraint.
\subsubsection{Expressive Preferences}
($\sim$) The utility function can support expressivity and context sensitivity, but is limited by only being defined over one type of constraint.
\subsubsection{Action Affordances}
($\times$) Is not sensitive to action affordances at runtime.
\subsubsection{Situation Model}
($\sim$) Updates a belief state based on observations, but will not update its world model expectations based on novel information.
\subsubsection{Conflict Structure}
($\times$) Resolves only a single type of conflict.
\subsubsection{Information Quality}
($\sim$) Will gather information about the state, but only akin to observational certainty for predefined features.
\subsubsection{Mitigation Utility}
($\sim$) It does assess the utility of information gathering.

\subsection{\citet{loreggia_making_2022}} They create AI agents modeling human decision making of when to break a rule according to multi-alternative decision field theory. But, the method depends on pre-training to match human preferences in a domain.
\subsubsection{Constraints \& Frames}
($\sim$) Integrates multiple types of constraints, but lacks reframing.
\subsubsection{Expressive Preferences}
($\sim$) Their implementation of MDFT enables predefined tradeoffs among predefined features.
\subsubsection{Action Affordances}
($\times$) The have a decision-making procedure that does not select or compose based on affordances at runtime.
\subsubsection{Situation Model}
($\sim$) Takes situational features as inputs, but does not update its model.
\subsubsection{Conflict Structure}
($\times$) Does not determine multiple types of conflicts.
\subsubsection{Information Quality}
($\times$) The model assumes reliable inputs.
\subsubsection{Mitigation Utility}
($\times$) The MDFT tradeoffs are not based on a notion of resolution, nor sensitive to conflict structure.

\subsection{\citet{simpkins_composable_2019}} The use of a high-level arbitrator policy to determine how to respond can theoretically (depending on the arbitrator policy) provide online conflict response capability sensitive to the situation with complex preferences. However, it is not clear how to extend the policy online when the agent's conflict mitigation implicit in the arbitrator's selection needs adaptation and/or existing modules lack an appropriate response (such as for constraint reframing).
\subsubsection{Constraints \& Frames}
($\sim$) There is potentially support for mulitple frames among multiple policies, and selection of a policy could potentially constitute reframing, but when comparing to the piracy interdiction scenario, it is unclear how this can be expected in a novel situation to frame a constraint that was insufficiently framed for a conflict.
\subsubsection{Expressive Preferences}
(\checkmark) A sufficiently expressive arbitrator policy could be sensitive to grounding states and represent complex preferences over which policy (mitigation response) to use.
\subsubsection{Action Affordances}
($\sim$) While it can flexibly select actions, it does not support runtime sensitivity to how an affordance relates to a novel conflict situation. 
\subsubsection{Situation Model}
($\sim$) The arbitrator is conditioned on the state, but it is unclear how it could update expectations in its world model.
\subsubsection{Conflict Structure}
($\sim$) While an arbitrator could be trained to react differently based on conflict structure, it is unclear how it would apply that assessment to novel conflicts for other processing to condition on.
\subsubsection{Information Quality}
($\times$) The arbitrator can't assess the state information in a module to assess quality.
\subsubsection{Mitigation Utility}
($\times$) Especially in a novel conflict, it is unclear how the arbitrator would predicate a response online based on the expectation of a particular type of conflict being resolved.

\subsection{\citet{choi_resolving_2025}} A formally-verified safety layer can interdict when an intended action is unsafe to provide an online response to imminent constraint conflict, but this work focuses on avoiding violations. It does not represent preferences for how to behave during unavoidable conflict.
\subsubsection{Constraints \& Frames}
($\times$) Performance (goals) are distinct from safety constraints and are handled differently. However, there really is only one type of constraint (collision) considered.
\subsubsection{Expressive Preferences}
($\times$) Safety always wins.
\subsubsection{Action Affordances}
($\sim$) The safety layer is a pretrained override mechanism selecting one kind of response action (avoidance) based on an agent's affordances, but will not enable novel uses of actions for untrained conflicts.
\subsubsection{Situation Model}
($\times$) The safety layer is effectively checking the state against potential violations, but does not support online world model updating.
\subsubsection{Conflict Structure}
($\times$) The conflicts only are between safety and performance.
\subsubsection{Information Quality}
($\times$) The safety layer trusts its inputs, implicitly.
\subsubsection{Mitigation Utility}
($\sim$) The system selects actions based on its ability to avoid a conflict, but is limited by its lack of conflict structure knowledge.

\subsection{\citet{banerjee_dynamic_2024}} Dynamic Model Predictive Shielding provides a framework to monitor the safety of an incomplete policy, enabling an online planner to compute a new safe action if a violation is predicted. This provides online violation detection and situation-specific response. But, this method only supports one constraint frame (safety) and does not have an ability to select among violations.
\subsubsection{Constraints \& Frames}
($\times$) While multiple forms of obstacle avoidance are considered, they are one kind of constraint with respect to human constraint frames.
\subsubsection{Expressive Preferences}
($\times$) This method assumes that safety constraints always win.
\subsubsection{Action Affordances}
(\checkmark) The recovery mechanism is a planner that dynamically composes actions to resolve a conflict.
\subsubsection{Situation Model}
($\sim$) The situation model includes a predictive model for near-term expectations, but is not updated
\subsubsection{Conflict Structure}
($\times$) The conflicts are between safety and performance.
\subsubsection{Information Quality}
($\times$) It does not include reasoning about sensor quality or reliability, or more complex notions of information quality.
\subsubsection{Mitigation Utility}
($\sim$) There are not multiple recovery strategies in response to a conflict, but it does plan to mitigate the anticipated conflict.

\subsection{\citet{mclaren_computational_2006}} The use of case-based reasoning in ethical dilemmas, using reasoning from a similar past dilemma. This is discussed in the main text.
\subsubsection{Constraints \& Frames}
(\checkmark) This method can effectively reframe a constraint if an appropriate case is retrieved.
\subsubsection{Expressive Preferences}
(\checkmark) Retrieval of a case according to features in the situation, and a case base with relevant cases of sufficient expressivity, can enable this.
\subsubsection{Action Affordances}
($\sim$) the adaptation is limited by the structure of the retrieved case and the retrieval is limited by the query mechanism.
\subsubsection{Situation Model}
($\times$) The current situation is the basis for a query, but there is no predictive model of the situation that is maintained and updated.
\subsubsection{Conflict Structure}
($\sim$) Implicitly, conflict structure may be captured by the query mechanism.
\subsubsection{Information Quality}
($\times$) There is no mechanism for assessing the quality of novel external information.
\subsubsection{Mitigation Utility}
($\sim$) The retrieval of a case cannot be assured to provide a response expected to resolve the conflict based on the structure of the current conflict.

\subsection{\citet{aha_goal_2018}} This work refers to goal reasoning agents that can manage their own goals with varied goal selection strategies and is discussed in the main text.
\subsubsection{Constraints \& Frames}
($\sim$) Different types of goals can be likened to different kinds of constraints, but they do not characterize an ability to reframe goals consistent with different types of human constraints.
\subsubsection{Expressive Preferences}
($\sim$) The selection between goals is not typically presented as sensitive to the kinds of features present in constraints, but this mechanism can potentially be updated.
\subsubsection{Action Affordances}
($\sim$) Policies are often linked to goals, where changing a goal is usually a strategy switch, not enabling flexible transfer.
\subsubsection{Situation Model}
(\checkmark) The basis for goal updating is expectation failure, depending on a predictive model of the situation, and one that is repaired through some goals.
\subsubsection{Conflict Structure}
($\times$)  While expectation failures are recognized, conflicts are not characterized by their structure.
\subsubsection{Information Quality}
(\checkmark) Both a mechanism for explaining discrepancies and the ability to select measurement goals provide information quality knowledge.
\subsubsection{Mitigation Utility}
($\sim$) Goal lifecycle strategies provide some mechanism for selecting between responses to a conflict, but are not clearly sensitive to conflict structure and belief in the mitigation of a conflict.

%% file: main.bbl
\begin{thebibliography}{39}
\providecommand{\natexlab}[1]{#1}

\bibitem[{Aha(2018)}]{aha_goal_2018}
Aha, D.~W. 2018.
\newblock Goal {Reasoning}: {Foundations}, {Emerging} {Applications}, and {Prospects}.
\newblock \emph{AI Magazine}, 39(2): 3--24.
\newblock Publisher: Wiley.

\bibitem[{Awad et~al.(2024)Awad, Levine, Loreggia, Mattei, Rahwan, Rossi, Talamadupula, Tenenbaum, and Kleiman-Weiner}]{awad_when_2024}
Awad, E.; Levine, S.; Loreggia, A.; Mattei, N.; Rahwan, I.; Rossi, F.; Talamadupula, K.; Tenenbaum, J.; and Kleiman-Weiner, M. 2024.
\newblock When is it acceptable to break the rules? {Knowledge} representation of moral judgements based on empirical data.
\newblock \emph{Autonomous Agents and Multi-Agent Systems}, 38(2): 35.

\bibitem[{Banerjee et~al.(2024)Banerjee, Rahmani, Biswas, and Dillig}]{banerjee_dynamic_2024}
Banerjee, A.; Rahmani, K.; Biswas, J.; and Dillig, I. 2024.
\newblock Dynamic {Model} {Predictive} {Shielding} for {Provably} {Safe} {Reinforcement} {Learning}.
\newblock In Globerson, A.; Mackey, L.; Belgrave, D.; Fan, A.; Paquet, U.; Tomczak, J.; and Zhang, C., eds., \emph{Advances in {Neural} {Information} {Processing} {Systems}}, volume~37, 100131--100159. Curran Associates, Inc.

\bibitem[{Bratman(1987)}]{bratman_intentions_1987}
Bratman, M.~E. 1987.
\newblock \emph{Intentions, {Plans}, and {Practical} {Reason}}.
\newblock Cambridge, MA: Harvard University Press.

\bibitem[{Briggs and Scheutz(2015)}]{briggs_sorry_2015}
Briggs, G.~M.; and Scheutz, M. 2015.
\newblock "{Sorry}, {I} {Can}'t {Do} {That}": {Developing} {Mechanisms} to {Appropriately} {Reject} {Directives} in {Human}-{Robot} {Interactions}.
\newblock In \emph{{AAAI} {Fall} {Symposia}}, 32--36.

\bibitem[{Brutzman, Blais, and Hsin-Fu(2020)}]{brutzman_ethical_2020}
Brutzman, D.~P.; Blais, C.~L.; and Hsin-Fu, W. 2020.
\newblock Ethical {Control} of {Unmanned} {Systems}: {Lifesaving}/{Lethal} {Scenarios} for {Naval} {Operations}.
\newblock Technical {Report} NPS-USW-2020-001, Naval Postgraduate School.

\bibitem[{Censi et~al.(2019)Censi, Slutsky, Wongpiromsarn, Yershov, Pendleton, Fu, and Frazzoli}]{censi_liability_2019}
Censi, A.; Slutsky, K.; Wongpiromsarn, T.; Yershov, D.; Pendleton, S.; Fu, J.; and Frazzoli, E. 2019.
\newblock Liability, {Ethics}, and {Culture}-{Aware} {Behavior} {Specification} using {Rulebooks}.
\newblock In \emph{2019 {International} {Conference} on {Robotics} and {Automation} ({ICRA})}, 8536--8542. Montreal, QC, Canada: IEEE.

\bibitem[{Choi et~al.(2025)Choi, Aloor, Li, Mendoza, Balakrishnan, and Tomlin}]{choi_resolving_2025}
Choi, J.~J.; Aloor, J.~J.; Li, J.; Mendoza, M.~G.; Balakrishnan, H.; and Tomlin, C.~J. 2025.
\newblock Resolving {Conflicting} {Constraints} in {Multi}-{Agent} {Reinforcement} {Learning} with {Layered} {Safety}.
\newblock Version Number: 1.

\bibitem[{Floyd, Drinkwater, and Aha(2015)}]{floyd_trust-guided_2015}
Floyd, M.~W.; Drinkwater, M.; and Aha, D.~W. 2015.
\newblock Trust-guided behavior adaptation using case-based reasoning.
\newblock In \emph{Proceedings of the 24th {International} {Conference} on {Artificial} {Intelligence}}, 4261--4267.

\bibitem[{Freuder and Wallace(1992)}]{freuder_partial_1992}
Freuder, E.~C.; and Wallace, R.~J. 1992.
\newblock Partial constraint satisfaction.
\newblock \emph{Artificial Intelligence}, 58(1-3): 21--70.
\newblock Publisher: Elsevier.

\bibitem[{Goel et~al.(2024)Goel, Wei, Lymperopoulos, Churá, Scheutz, and Sinapov}]{goel_novelgym_2024}
Goel, S.; Wei, Y.; Lymperopoulos, P.; Churá, K.; Scheutz, M.; and Sinapov, J. 2024.
\newblock {NovelGym}: {A} {Flexible} {Ecosystem} for {Hybrid} {Planning} and {Learning} {Agents} {Designed} for {Open} {Worlds}.
\newblock In \emph{Proceedings of the 23rd {International} {Conference} on {Autonomous} {Agents} and {Multiagent} {Systems}}, {AAMAS} '24, 688--696. International Foundation for Autonomous Agents and Multiagent Systems.

\bibitem[{Haider et~al.(2024)Haider, Roscher, Herd, Schmoeller~Roza, and Burton}]{haider_can_2024}
Haider, T.; Roscher, K.; Herd, B.; Schmoeller~Roza, F.; and Burton, S. 2024.
\newblock Can you trust your {Agent}? {The} {Effect} of {Out}-of-{Distribution} {Detection} on the {Safety} of {Reinforcement} {Learning} {Systems}.
\newblock In \emph{Proceedings of the 39th {ACM}/{SIGAPP} {Symposium} on {Applied} {Computing}}, 1569--1578. Avila Spain: ACM.
\newblock ISBN 979-8-4007-0243-3.

\bibitem[{{ICC International Maritime Bureau}(2024)}]{icc_international_maritime_bureau_piracy_2024}
{ICC International Maritime Bureau}. 2024.
\newblock Piracy and {Armed} {Robbery} {Against} {Ships}: {Report} for the {Period} 1 {January} - 31 {December} 2023.
\newblock Technical report, ICC International Maritime Bureau (IMB), London, UK.

\bibitem[{Jenni and Loewenstein(1997)}]{jenni_explaining_1997}
Jenni, K.; and Loewenstein, G. 1997.
\newblock Explaining the identifiable victim effect.
\newblock \emph{Journal of Risk and Uncertainty}, 14(3): 235--257.

\bibitem[{Johnson et~al.(2018)Johnson, Floyd, Coman, Wilson, and Aha}]{johnson_goal_2018_different}
Johnson, B.; Floyd, M.~W.; Coman, A.; Wilson, M.~A.; and Aha, D.~W. 2018.
\newblock Goal {Reasoning} and {Trusted} {Autonomy}.
\newblock In \emph{Studies in {Systems}, {Decision} and {Control}}, 47--66. Cham: Springer International Publishing.
\newblock ISBN 978-3-319-64815-6 978-3-319-64816-3.

\bibitem[{Jones and Wray(2024)}]{jones_toward_2024_correct}
Jones, S.~J.; and Wray, R.~E. 2024.
\newblock Toward {Constraint} {Compliant} {Goal} {Formulation} and {Planning}.
\newblock In \emph{Proceedings of the {Advances} in {Cognitive} {Systems} {Conference}}. Palermo.

\bibitem[{Kaelbling, Littman, and Cassandra(1998)}]{kaelbling_planning_1998}
Kaelbling, L.~P.; Littman, M.~L.; and Cassandra, A.~R. 1998.
\newblock Planning and acting in partially observable stochastic domains.
\newblock \emph{Artificial intelligence}, 101(1-2): 99--134.
\newblock Publisher: Elsevier.

\bibitem[{Kejriwal et~al.(2024)Kejriwal, Kildebeck, Steininger, and Shrivastava}]{kejriwal_challenges_2024}
Kejriwal, M.; Kildebeck, E.; Steininger, R.; and Shrivastava, A. 2024.
\newblock Challenges, evaluation and opportunities for open-world learning.
\newblock \emph{Nat Mach Intell}, 6(6): 580--588.

\bibitem[{Klein(1998)}]{klein_sources_1998}
Klein. 1998.
\newblock \emph{Sources of {Power}}.
\newblock The MIT Press.

\bibitem[{Kuipers(2018)}]{kuipers_how_2018}
Kuipers, B. 2018.
\newblock How {Can} {We} {Trust} a {Robot}?
\newblock \emph{Communications of the ACM}, 61(3): 86--95.

\bibitem[{Langosco et~al.(2022)Langosco, Koch, Sharkey, Pfau, Orseau, and Krueger}]{langosco_goal_2022}
Langosco, L.; Koch, J.; Sharkey, L.; Pfau, J.; Orseau, L.; and Krueger, D. 2022.
\newblock Goal {Misgeneralization} in {Deep} {Reinforcement} {Learning}.
\newblock In \emph{{ICML} 2022}.
\newblock ArXiv:2105.14111 [cs].

\bibitem[{Leontarakis(2015)}]{leontarakis_piracy_2015}
Leontarakis, A.-A.~G. 2015.
\newblock \emph{Piracy, {Law} {Efficiency}, {Statistics} and {Possible} {Solutions} to that {Severe} and {Growing} {Threat}}.
\newblock Master's thesis, Hellenic Coast Guard Academy.

\bibitem[{Loreggia et~al.(2022)Loreggia, Mattei, Rahgooy, Rossi, Srivastava, and Venable}]{loreggia_making_2022}
Loreggia, A.; Mattei, N.; Rahgooy, T.; Rossi, F.; Srivastava, B.; and Venable, K.~B. 2022.
\newblock Making {Human}-{Like} {Moral} {Decisions}.
\newblock In \emph{Proceedings of the 2022 {AAAI}/{ACM} {Conference} on {AI}, {Ethics}, and {Society}}, 447--454. Oxford United Kingdom: ACM.
\newblock ISBN 978-1-4503-9247-1.

\bibitem[{Lu et~al.(2024)Lu, Svegliato, Nashed, Zilberstein, and Russell}]{lu_ethically_2024}
Lu, Q.; Svegliato, J.; Nashed, S.~B.; Zilberstein, S.; and Russell, S. 2024.
\newblock Ethically {Compliant} {Autonomous} {Systems} under {Partial} {Observability}.
\newblock In \emph{2024 {IEEE} {International} {Conference} on {Robotics} and {Automation} ({ICRA})}, 16229--16235. Yokohama, Japan: IEEE.

\bibitem[{McKie and Richardson(2003)}]{mckie_rule_2003}
McKie, J.; and Richardson, J. 2003.
\newblock The {Rule} of {Rescue}.
\newblock \emph{Social Science \& Medicine}, 56(12): 2407--2419.
\newblock Publisher: Elsevier BV.

\bibitem[{McLaren(2006)}]{mclaren_computational_2006}
McLaren, B. 2006.
\newblock Computational {Models} of {Ethical} {Reasoning}: {Challenges}, {Initial} {Steps}, and {Future} {Directions}.
\newblock \emph{IEEE Intelligent Systems}, 21(4): 29--37.
\newblock Publisher: Institute of Electrical and Electronics Engineers (IEEE).

\bibitem[{Meseguer, Rossi, and Schiex(2006)}]{rossi_soft_2006}
Meseguer, P.; Rossi, F.; and Schiex, T. 2006.
\newblock Soft {Constraints}.
\newblock In Rossi, F.; Van~Beek, P.; and Walsh, T., eds., \emph{Handbook of constraint programming}, Foundations of artificial intelligence. Amsterdam ; Boston: Elsevier, 1st ed edition.
\newblock ISBN 978-0-444-52726-4.
\newblock OCLC: ocm70408044.

\bibitem[{Milli et~al.(2017)Milli, Hadfield-Menell, Dragan, and Russell}]{milli_should_2017}
Milli, S.; Hadfield-Menell, D.; Dragan, A.; and Russell, S. 2017.
\newblock Should robots be obedient?
\newblock In \emph{Proceedings of the 26th {International} {Joint} {Conference} on {Artificial} {Intelligence}}, {IJCAI}'17, 4754--4760. Melbourne, Australia: AAAI Press.
\newblock ISBN 978-0-9992411-0-3.

\bibitem[{Mohan et~al.(2024)Mohan, Piotrowski, Stern, Grover, Kim, Le, Sher, and De~Kleer}]{mohan_domain-independent_2024}
Mohan, S.; Piotrowski, W.; Stern, R.; Grover, S.; Kim, S.; Le, J.; Sher, Y.; and De~Kleer, J. 2024.
\newblock A domain-independent agent architecture for adaptive operation in evolving open worlds.
\newblock \emph{Artificial Intelligence}, 334: 104161.

\bibitem[{Molineaux, Klenk, and Aha(2010)}]{molineaux_goal-driven_2010}
Molineaux, M.; Klenk, M.; and Aha, D. 2010.
\newblock Goal-driven autonomy in a {Navy} strategy simulation.
\newblock In \emph{Proceedings of the {AAAI} {Conference} on {Artificial} {Intelligence}}, volume~24, 1548--1554.
\newblock Issue: 1.

\bibitem[{Newell(1969)}]{newell_heuristic_1969}
Newell, A. 1969.
\newblock Heuristic {Programming}: {Ill}-structured {Problems}.
\newblock In Aronofsky, J., ed., \emph{Progress in {Operations} {Research} {III}}, 360--414. New York: Wiley.

\bibitem[{Newell(1982)}]{newell_knowledge_1982}
Newell, A. 1982.
\newblock The knowledge level.
\newblock \emph{Artificial Intelligence}, 18(1): 82--127.

\bibitem[{Payne, Bettman, and Johnson(1993)}]{payne_adaptive_1993}
Payne, J.~W.; Bettman, J.~R.; and Johnson, E.~J. 1993.
\newblock \emph{The {Adaptive} {Decision} {Maker}}.
\newblock Cambridge: Cambridge University Press.
\newblock ISBN 978-0-521-41505-7.

\bibitem[{Rao and Georgeff(1995)}]{rao_bdi_1995}
Rao, A.~S.; and Georgeff, M.~P. 1995.
\newblock {BDI} {Agents}: {From} {Theory} to {Practice}.
\newblock In \emph{Proceedings of the {First} {International} {Conference} on {Multiagent} {Systems}}.

\bibitem[{Shah et~al.(2019)Shah, Krasheninnikov, Alexander, Abbeel, and Dragan}]{shah_preferences_2019}
Shah, R.; Krasheninnikov, D.; Alexander, J.; Abbeel, P.; and Dragan, A. 2019.
\newblock Preferences {Implicit} in the {State} of the {World}.
\newblock In \emph{{ICLR} 2019}. arXiv.
\newblock ArXiv:1902.04198 [cs].

\bibitem[{Simon(1973)}]{simon_structure_1973}
Simon, H.~A. 1973.
\newblock The {Structure} of {Ill} {Structured} {Problems}.
\newblock \emph{Artificial Intelligence}, 4(3-4): 181--201.

\bibitem[{Simpkins and Isbell(2019)}]{simpkins_composable_2019}
Simpkins, C.; and Isbell, C. 2019.
\newblock Composable {Modular} {Reinforcement} {Learning}.
\newblock \emph{Proceedings of the AAAI Conference on Artificial Intelligence}, 33(01): 4975--4982.
\newblock Publisher: Association for the Advancement of Artificial Intelligence (AAAI).

\bibitem[{Smith(2004)}]{smith_choosing_2004}
Smith, D.~E. 2004.
\newblock Choosing objectives in over-subscription planning.
\newblock In \emph{Proceedings of the {Fourteenth} {International} {Conference} on {International} {Conference} on {Automated} {Planning} and {Scheduling}}, 393--401.

\bibitem[{Wray, Jones, and Laird(2023)}]{wray_computational-level_2023}
Wray, R.~E.; Jones, S.; and Laird, J.~E. 2023.
\newblock Computational-level {Analysis} of {Constraint} {Compliance} for {General} {Intelligence}.
\newblock In \emph{Proc. of {Artificial} {General} {Intelligence} ({AGI}) {Conf}.} Stockholm.

\end{thebibliography}
